\RequirePackage{fix-cm}
\documentclass[twocolumn]{svjour3}          % twocolumn
\smartqed  % flush right qed marks, e.g. at end of proof
\usepackage{graphicx}
\usepackage{amssymb}
\usepackage[cmex10]{amsmath}
\usepackage{array}
\usepackage{url}
\usepackage{color}
\usepackage[tight,footnotesize]{subfigure}
\usepackage{natbib}
\usepackage{hyphenat}
\usepackage[pdftex,
CJKbookmarks=true, bookmarksnumbered=true, bookmarksopen=true,
colorlinks=true,
pdfborder=001,
citecolor=blue, linkcolor=blue, anchorcolor=red, urlcolor=blue
]{hyperref}

\usepackage[title]{appendix}%
\usepackage{bbm,dsfont}
\usepackage{indentfirst} 
\usepackage{booktabs}

\usepackage[dvipsnames]{xcolor} % Text color
\usepackage{caption}
\usepackage{subcaption}
\usepackage{multirow}
\usepackage{tabularx}
\usepackage{adjustbox}
\usepackage{pifont}
\usepackage[skip=2pt]{caption}
\usepackage{marvosym}

\usepackage{bm}
\usepackage{mdwmath}
\usepackage{mdwtab}
\usepackage{eqparbox}
\usepackage{multirow}
\usepackage{amsfonts}
\usepackage{epsfig}
\usepackage{booktabs}
\usepackage{algorithm,algorithmic}
\usepackage{amssymb}

\renewcommand{\arraystretch}{1.3}

\definecolor{darkmag}{rgb}{0.55,0,0.55}
\begin{document}
\begin{sloppypar}

%\title{ Dynamic Contrastive Reconstruction Loss for Image Restoration in Adverse Weather Conditions
%Robust Consistent and Dynamic Contrast-Assisted Network for Image Restoration in Adverse Weather Conditions%\thanks{Grants or other notes
% Progressive Negative Enhancing Contrastive Learning for Adverse Weather Removal
\title{Towards Generalized Proactive Defense against Face Swapping with Contour-Hybrid Watermark}

\titlerunning{Towards Generalized Proactive Defense against Face Swapping with Contour-Hybrid Watermark}        % if too long for running head

% \author{Lingfeng He \and
%         De Cheng\textsuperscript{\Letter} \and
%         Nannan Wang\textsuperscript{\Letter} \and
%         Xinbo Gao \and
% }

\author{Ruiyang Xia \and 
        Dawei Zhou \and 
        Decheng Liu \and 
        Lin Yuan \and
        Jie Li \and
        Nannan Wang\and
        Xinbo Gao}

\authorrunning{Ruiyang Xia et al} % if too long for running head

\institute{
 Ruiyang Xia \at
 Xidian University, Xi'an 710071, Shaanxi, China \\
\email{\url{ryon@stu.xidian.edu.cn}} \\
 \\Dawei Zhou \at
 City University of Macau, Macao Special Administrative Region, China \\
\email{\url{dwzhou.xidian@gmail.com}} \\
 \\Decheng Liu \at
 Xidian University, Xi'an 710071, Shaanxi, China \\
\email{\url{dchliu@xidian.edu.cn}} \\
 \\Lin Yuan \at
 Chongqing University of Posts and Telecommunications, Chongqing 400065, China \\
\email{\url{yuanlin@cqupt.edu.cn}} \\
 \\Jie Li \at
 Xidian University, Xi'an 710071, Shaanxi, China \\
\email{\url{leejie@mail.xidian.edu.cn}} \\
 \\Nannan Wang \at
 Xidian University, Xi'an 710071, Shaanxi, China \\
\email{\url{nnwang@xidian.edu.cn}} \\
 \\Xinbo Gao \at
 Xidian University, Xi'an 710071, Shaanxi, China\\
\email{\url{xbgao@mail.xidian.edu.cn}} \\
}

\date{Received: date / Accepted: date}
% % The correct dates will be entered by the editor
\maketitle

\begin{abstract}
Face swapping, recognized as a privacy and security concern, has prompted considerable defensive research. With the advancements in AI-generated content, the discrepancies between the real and swapped faces have become nuanced. Considering the difficulty of forged traces detection, we shift the focus to the face swapping purpose and proactively embed elaborate watermarks against unknown face swapping techniques. Given that the constant purpose is to swap the original face identity while preserving the background, we concentrate on the regions surrounding the face to ensure robust watermark generation, while embedding the contour texture and face identity information to achieve progressive image determination. The watermark is located in the facial contour and contains hybrid messages, dubbed the \underline{\textbf{c}}ontour-hybrid water\underline{\textbf{mark}} (CMark). Our approach generalizes face swapping detection without requiring any swapping techniques during training and the storage of large-scale messages in advance. Experiments conducted across 8 face swapping techniques demonstrate the superiority of our approach compared with state-of-the-art passive and proactive detectors while achieving a favorable balance between the image quality and watermark robustness. %Code is available at \url{https://github.com/xarryon/CMark}.
\end{abstract}

\section{Introduction}
\noindent With the development of AI-generated content (AIGC) towards high fidelity and low complexity \cite{aigc}, it is convenient to generate fictitious but vivid media. However, this convenience raises the forgery dissemination, especially for face swapping techniques. Recent reports reveal a 704\% surge in the use of face swapping techniques in 2023 to bypass identity verification, underscoring face swapping as the most prevalent deepfake \cite{report}. To make AI trustworthy \cite{trust,trust2}, face swapping detection is naturally introduced \cite{ddsurvey}.
\begin{figure}
\centering
\includegraphics[width=\linewidth]{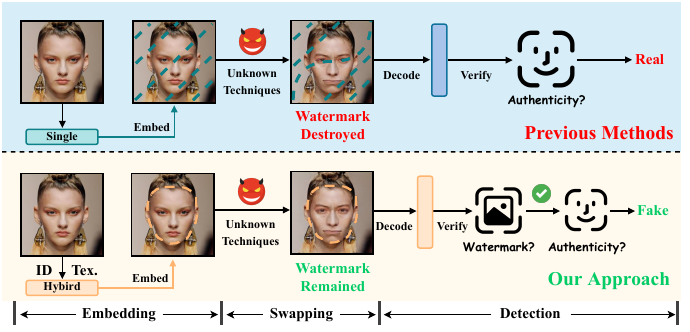}
\caption{\textbf{Comparison of proactive detection.} Previous methods embed a watermark with a single-type message into the entire image. However, we embed the watermark into the facial contour to ensure robustness, while integrating identity and contour texture (Tex.) into the message to achieve progressive determination.}
\label{f1}
\end{figure}
\par Previous passive detectors mainly focus on the pixel discrepancies between the real and swapped images \cite{fsd,caddm,fpg,liu,prodet,oddn,DATA}. However, their detection performance is constrained by diverse forged traces and continuously decreased discrepancies \cite{faceswap_survey}. Proactive paradigms, which embed watermarks in images and trace them after forgery, have been proposed to improve detection accuracy and make determined evidence more intuitive \cite{robust,sepmark,proactive,lampmark}.
\par Conventionally, as shown in Fig.~\ref{f1}, a single-type message, such as random Gaussian variables \cite{faketagger,facesigns,sepmark}, is allocated to each image. The watermark is then generated and applied across the entire image. Although the robustness of the watermark remains when training with specific face swapping techniques \cite{facesigns,sepmark,robust,lampmark}, according to the reported bit error rates from \cite{lampmark}, some proactive detectors become sensitive when exposed to unknown techniques, thereby resulting in poor generalization. Furthermore, this type of message only ensures the determination of the images that have already been incorporated into the watermarks. However, the distinction between the watermarked and non-watermarked images is ignored, which limits the applicability of the proactive detection methods in the real-world as there are many non-watermarked images spread on the websites. Therefore, a generalized and applicable proactive approach is needed imminently.
\par Starting from the constant face swapping purpose, \textit{i.e.}, \textbf{swapping the original face identity while preserving the background}. A novel proactive detection approach dubbed \underline{\textbf{c}}ontour-hybrid water\underline{\textbf{mark}} (\textbf{CMark}) is proposed to effectively defend against unknown face swapping techniques. Unlike previous works that embed a single-type message into an entire image \cite{zhu2018hidden,pimog,arwgan,sepmark}, the watermark location and message are carefully designed in Fig.~\ref{f1} for robust watermark generation and progressive image determination. As to the location, we focus on the contour region, which reduces the interferences brought by face swapping and malicious background cropping during watermark decoding. Moreover, the local watermark facilitates the visual quality as most regions of the watermarked images are consistent with the originals. Given that the similar contour but different identity for real and swapped images, we integrate the contour texture and face identity information as the messages. Consequently, the verification is separated into two steps. The message involving contour texture is initially verified to determine whether the image is watermarked, while the decoded identity message is subsequently verified for authenticity determination. Different from relevant methods \cite{robust,facesigns,proactive,dual}, our approach is self-verifiable \cite{lampmark}, which verifies the decoded messages without requiring the corresponding preset messages, thereby efficiently saving the message storage. Before verification, messages are encrypted to ensure the security of the proposed proactive detector.
\par The main contributions can hence be summarized as:
\begin{itemize}
\item To make the proactive defense approach generalizes to unknown face swapping techniques, CMark is proposed based on the constant swapping purpose. To the best of our knowledge, we are the early exploration that simultaneously concentrates on designing the watermark location and message, which ensures the visual quality, watermark robustness, and progressive image determination.
\item Our approach achieves accurate detection without requiring any face swapping techniques during the model training. Furthermore, the self-verifiable property is maintained to save the storage of messages \textit{w.r.t.} all watermarked images. 
\item Extensive experiments across 8 prevalent face swapping techniques demonstrate the superiority of our approach compared with passive and proactive detectors. Moreover, the results regarding visual quality and watermark robustness underscore the favorable balance achieved by the CMark model.
\end{itemize}
\par The rest of this paper is organized as follows. Sect.~\ref{sec:2} briefly reviews the related work. Sect.~\ref{sec:method} lists the motivation and threat model as the prerequisite of the proposed approach. Sect.~\ref{sec:4} thus presents the detail of the proposed CMark framework. Sect.~\ref{sec:5} shows the experimental results and corresponding analysis. The limitations and future perspective are presented in Sect.~\ref{sec:6}. Finally, the conclusion is summarized in Sect.~\ref{sec:7}.
\section{Related Work}
\label{sec:2}
\subsection{Face Swapping Techniques}
\noindent Early research swap faces with manual techniques. Given a source and target image, The work in \cite{faceswap} decomposed the facial areas of the source and target images into Delaunay triangles. The source face is swapped by replacing its triangles with the corresponding wrapped ones from the target face. However, in cases with substantial discrepancies in facial attributes between the source and target faces, the results suffered from reduced fidelity. Deep learning is then introduced to elevate the fidelity of the generated swapped images. Specifically, reconstruction-based methods \cite{mobile,faceshifter,simswap} mainly focus on extracting source attributes and target identity through reconstruction learning before integration. Due to entanglement of features in attributes and identity, StyleGAN-based methods \cite{megafs,e4s} are proposed to disentangle face images into multiple representative latent vectors \cite{stylegan}, thereby achieving more fidelity face images with detailed texture preservation. Based on the powerful diffusion models \cite{ddim,ddpm,iddpm,ldm} that stabilize the training process and progressively fit the complex distribution, diffusion-based methods \cite{diffface,diffswap} have emerged to elevate the fidelity across varying scales. 
\par While various swapping techniques produce distinct forged traces, their overarching purpose remains consistent. This fundamental principle enables the establishment of a generalized face swapping detection model.
\subsection{Face Forgery Detection}
\noindent Prevalent face forgery detection can be categorized as model-based and data-based methods \cite{ddsurvey}. The former aims to improve the detection model awareness of forged traces within the swapped images. For instance, IID \cite{fsd} decomposed the swapped image into explicit and implicit embedding for detection. F3Net \cite{f3} simultaneously analyzed the face images from the spatial and frequency domain. RECCE \cite{recce} amplified the discrepancies between the real and swapped images through face reconstruction. To decrease the overfitting of the evident traces from the specific forgery techniques, data-based methods \cite{xray,sbi,fpg,freq,time} try to emphasize the common forged traces by simulating the color mismatching, face blurring, and blending inconsistency into the real images. For example, \cite{xray} simulate the face swapping process by deliberately incorporating inconsistency within the face edge. Considering the complexity of similar face searching between the source and target face, \cite{sbi} proposed a self-blending strategy and simulated more forged traces. Subsequently, \cite{fpg} strengthened the local hard forgery traces for training through the interaction with the detector perception. \cite{freq} and \cite{time} further expanded the self-blending strategy to the frequency and temporal domain, respectively.
\par Nevertheless, the forged traces generated from the swapped techniques are becoming imperceptible. Previous passive detectors may prove ineffective against recent advancements such as StyleGAN-based or diffusion-based methods. Proactive paradigms are thus introduced to embed watermarks and analyze the change of the watermark after forgery, which have been proposed to improve detection accuracy.

\subsection{Deep Image Watermarking}
\noindent Beyond visible watermarks, the pursuit of invisibility and robustness has become a primary focus in related \cite{zhu2018hidden,mbrs,arwgan,pimog}. \cite{zhu2018hidden} first proposed an end-to-end robust watermark network, which comprises an image encoder, a random noise layer, and a watermark decoder. Considering that JPEG is non-differential to the network, MBRS \cite{mbrs} randomly attached each sample within a batch with identical mapping, simulated JPEG, or real JPEG compression. PIMOG \cite{pimog} formulated the most influenced distortions occurring during the screen-capturing process within the noise pool. ARWGAN \cite{arwgan} integrated dense blocks and attention mechanism to enhance the image quality and watermark robustness. As the generated watermarked images are employed for copyright tracing, recent detection methods have increasingly focused on proactively embedding watermarks into images to monitor the variations of watermark \cite{proactive,facesigns,sepmark,lampmark,robust}. Facesigns \cite{facesigns} proposed a semi-fragile model to make the generated watermark sensitive to the forgery techniques. SepMark \cite{sepmark} developed robust and semi-fragile branches to assess their decoding consistency. IDMark \cite{proactive} introduced watermarked identity features and further correlated the features with preset watermarks to determine image authenticity. The work in \cite{robust} compared the watermark with the stored protected identity messages. \cite{lampmark} achieves self-verifiable by generating landmark watermark can compare it with the landmark from the forged images.
\par Due to the diversity of face swapping techniques, the presence of non-watermarked images, and the large volume of watermarked images, it poses significant challenges for existing methods in real-world applications.

\section{Motivation and Threat Model}
\label{sec:method}
\subsection{Motivation}
\begin{figure}
\centering
\includegraphics[width=\linewidth]{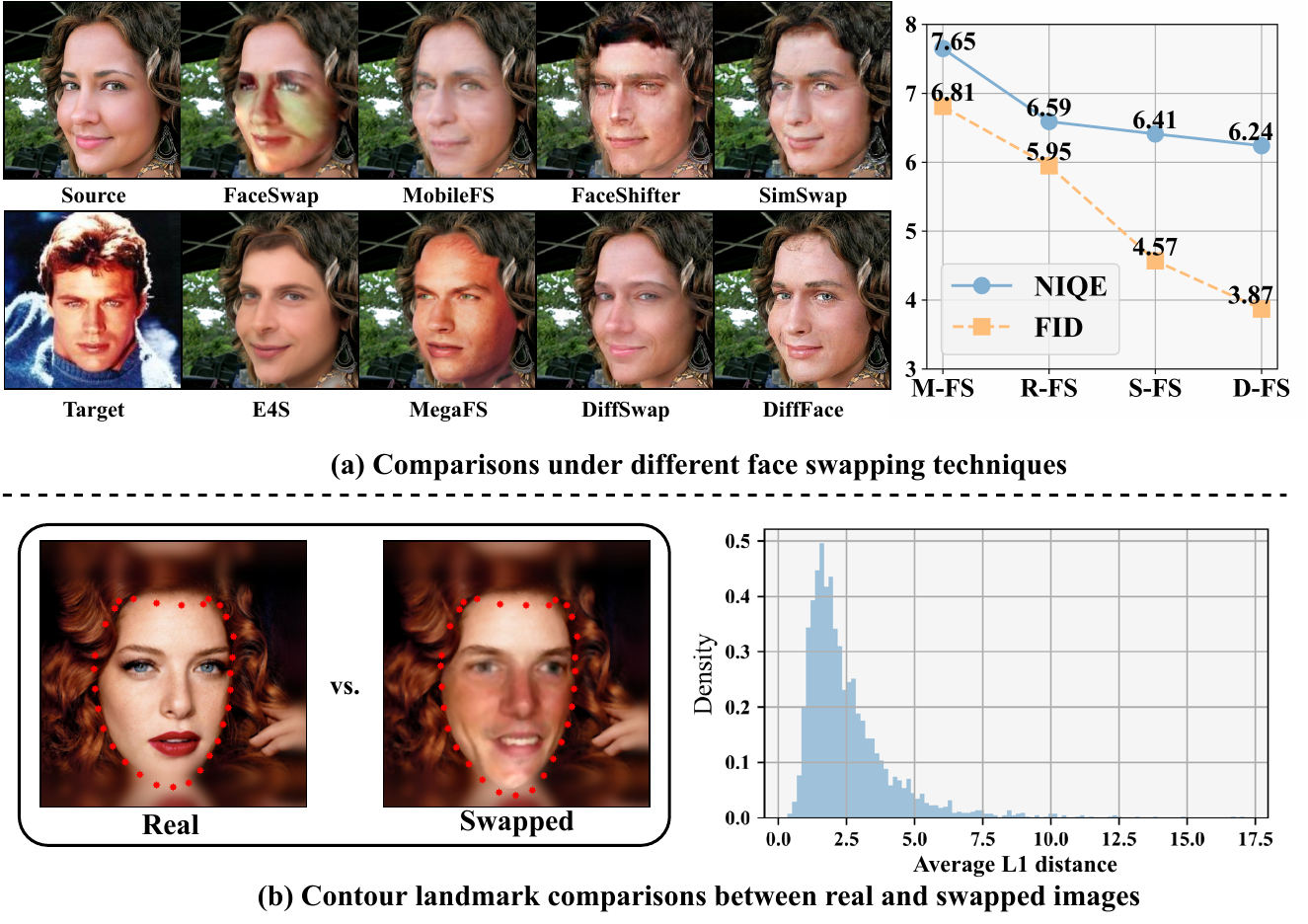}
\caption{\textbf{(a)} Qualitative and quantitative face swapping comparisons. `M-FS', `R-FS', `S-FS', and `D-FS' denote manual method (FaceSwap), reconstruction-based methods (MobileFS, Faceshifter, and SimSwap), StyleGAN-based methods (E4S and MegaFS), and Diffusion-based methods (DiffSwap and DiffFace). FID is computed by averaging the features from different layers of \cite{inception}. \textbf{(b)} Qualitative and quantitative contour landmark comparisons under FaceSwap \cite{faceswap} on 256$\times$256 resolution.}
\label{f2}
\end{figure}

\noindent\textbf{Diverse forged traces but constant purpose.} 
The swapped face images under different techniques are illustrated in Fig.~\ref{f2}(a), highlighting that the forged traces are inherently dependent on specific techniques. For example, FaceSwap \cite{faceswap} brings color mismatching, MobileFS \cite{mobile} and FaceShifter \cite{faceshifter} emerge artifacts within the face. Furthermore, the decreased NIQE \cite{niqe} and FID \cite{fid} scores indicate the continuously improved quality of the swapped images, which further increases the detection difficulties. While numerous techniques are unknown to defenders,  they share a common purpose: swapping the original identity while preserving the background. Furthermore, since face swapping mainly modifies the internal attributes, the qualitative result in Fig.~\ref{f2}(b) shows minimal deviations in contour landmarks between the real and swapped face. Quantitative analysis reveals that most landmark discrepancies between the paired images have less than 5 average L1 distance. These observations motivate a face-aware watermarking strategy, wherein the watermark distribution is controlled to ensure robustness and face swapping detection is achieved through message tailoring.

\noindent\textbf{Omission in previous proactive detectors.} Four problems that are necessary but less discussed in previous proactive methods: (1) Although the authenticity of watermarked images can be assessed, the determination between the watermarked and non-watermarked images is frequently neglected. (2) Proactive detection is typically achieved through training with the corresponding forgery techniques, which hinders the generalized detection due to the varying nature of forgery traces. (3) Certain proactive methods depend on matching preset messages stored in the memory, which imposes considerable strain on storage resources. (4) The robustness of the watermark remains inadequately addressed when face swapping techniques are combined with image distortions. As proactive detection continues to evolve, more considerations are essential to improve the practicality of the proactive detectors in real-world scenarios.
\subsection{Multiple Threatening Cases}
\noindent Given that the variety of situations in real-world scenarios, we outline three representative threatening cases based on a security assumption as follows:
\begin{itemize}
\item \textbf{Security assumption:} All models are deployed \textbf{securely} on clients, while message verification is conducted on a central platform, ensuring that the model parameters and the verified process are \textbf{forbidden} for attackers. Accessible things for them are images posted on websites and messages intercepted during client-platform communication.
\item \textbf{Case 1:} Before swapping the watermarked face images, attackers extract the corresponding backgrounds and intentionally crop some regions to erase the embedded watermarks.
\item \textbf{Case 2:} After swapping the watermarked face images, attackers impose different images distortions to interfere with the watermark decoding process.
\item \textbf{Case 3:} During the communication between clients and the platform, attackers surveil network traffic to intercept the unencrypted decoded messages.  Subsequently, they infiltrate the communication channel and introduce tampered messages into the verification process, thereby hindering the confidence.
\end{itemize}

\begin{figure*}
\centering
\includegraphics[width=\linewidth]{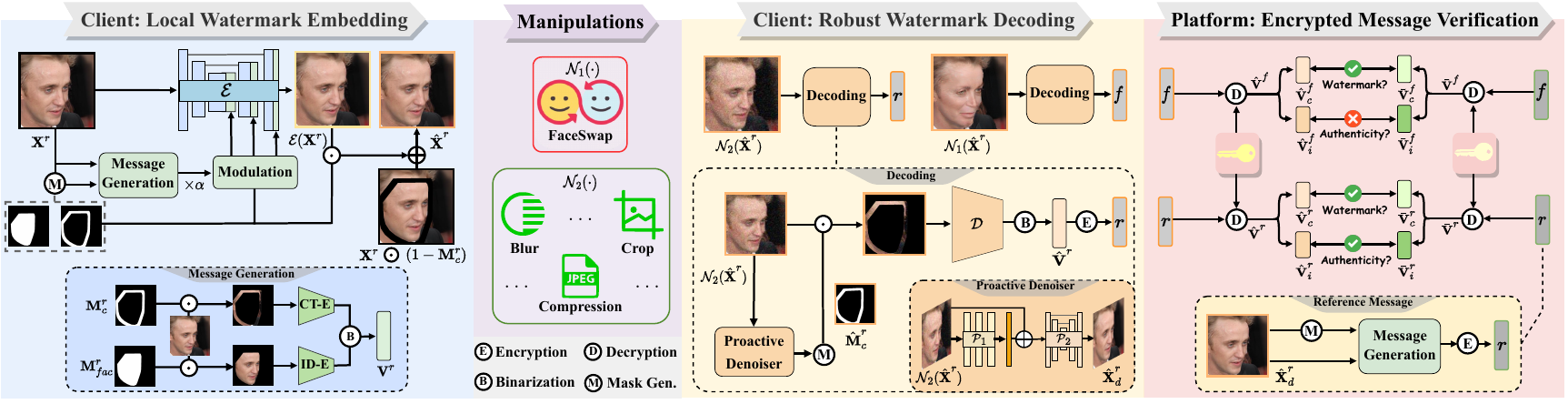}
\caption{\textbf{Pipeline of our proposed CMark model}. \textbf{Firstly}, the client generates message $\textbf{V}^r$ by integrating features from the contour texture extractor (CT-E) and face identity extractor (ID-E). The message is then embedded into the facial contour through watermark encoder $\mathcal{E}$ after scaling by a strength factor $\alpha$. \textbf{Secondly}, for the watermarked image $\hat{\textbf{X}}^{r}$ subjected to random manipulations, the robust watermark is decoded by inputting the contour region of the manipulated $\hat{\textbf{X}}^{r}$ into the watermark decoder $\mathcal{D}$. \textbf{Finally}, after sending the encrypted messages to the platform for decryption, the decoded message $\hat{\textbf{V}}$ is verified against the reference message $\bar{\textbf{V}}$ to determine the watermark existence and the image authenticity. `Gen.' denotes `Generation'.}
\label{f3}
\end{figure*}

\section{Contour-Hybrid Watermark}
\label{sec:4}
\subsection{Problem Formulation}\noindent Given a real image $\text{\textbf{X}}^{r}$$\in$$\mathbb{R}^{H\times W\times3}$ with identity message $\text{\textbf{V}}_{i}^{r}$$\in$$\{0,1\}^{1\times C}$, the image is decomposed into a face $\text{\textbf{R}}_{fac}^{r}$ and background region $\text{\textbf{R}}_{bac}^{r}$$\in$$\mathbb{R}^{H\times W\times3}$. Assuming that unknown face swapping techniques are employed to tamper the $\text{\textbf{X}}^{r}$ as $\text{\textbf{X}}^{f}$, both images meet $\text{\textbf{R}}_{bac}^{r}$$\sim$$\text{\textbf{R}}_{bac}^{f}$, $\text{\textbf{R}}_{fac}^{r}$$\nsim$$\text{\textbf{R}}_{fac}^{f}$, and $\text{\textbf{V}}_{i}^{r}$$\neq$$\text{\textbf{V}}_{i}^{f}$. Therefore, three stages are involved in the model to proactively defend against such techniques. As shown in Fig.~\ref{f3}, after extracting the contour mask $\text{\textbf{M}}_{c}^{r}$$\in$$\{0,1\}^{H\times W\times 3}$, a watermark is imperceptibly embedded within contour region $\text{\textbf{R}}_{c}^{r}$ via an encoder $\mathcal{E}$. When the watermarked image $\hat{\textbf{X}}^{r}$$\in$$\mathbb{R}^{H\times W\times3}$ is manipulated (either face swapping $\mathcal{N}_1(\cdot)$ or normal distortions $\mathcal{N}_2(\cdot)$), the robust watermark is decoded by $\mathcal{D}$ and transmitted to the platform for verification. The decoded message comprises the contour texture $\hat{\textbf{V}}_{c}^{r}$ and face identity message $\hat{\textbf{V}}_{i}^{r}$, which are verified with the reference messages $\bar{\textbf{V}}_{c}$ and $\bar{\textbf{V}}_{i}$ from the denoised image $\hat{\textbf{X}}_{d}$ to enable progressive determination. Based on the security assumption, the decoded and reference messages are encrypted before verification.

\subsection{Local Watermark Embedding}
\begin{figure}
\centering
\includegraphics[width=\linewidth]{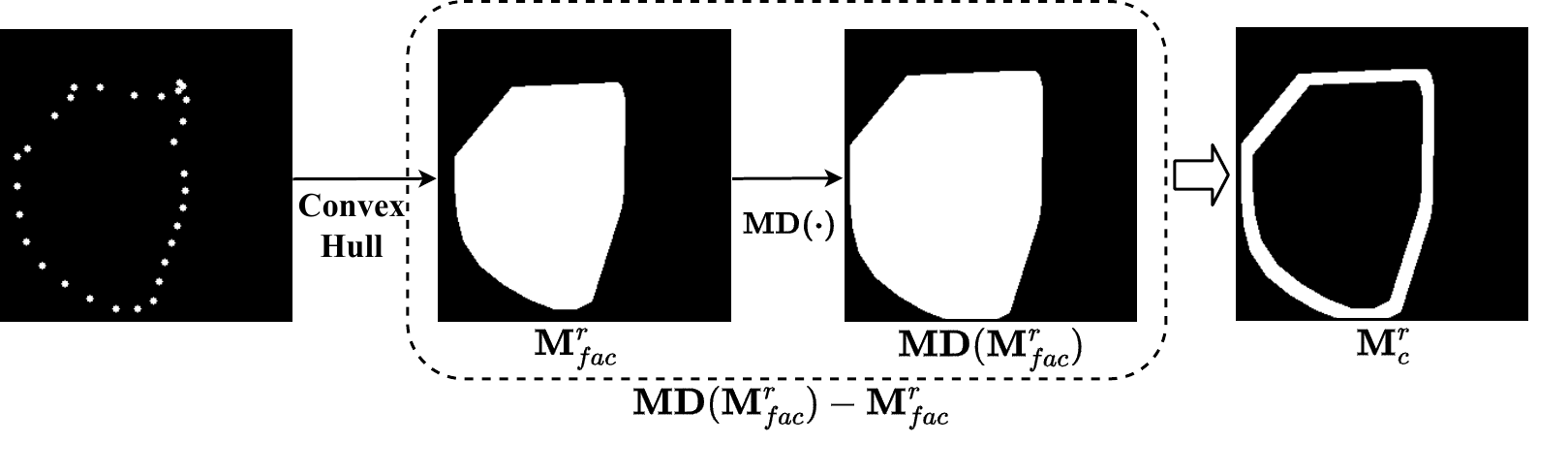}
\caption{\textbf{Illustration of the facial contour mask generation.} $\text{MD}(\cdot)$ denotes the morphological dilation function. This process confines the watermark to sub-regions of the background, rendering it unaffected by internal facial manipulations.}
\label{f4_new}
\end{figure}
\noindent\textbf{Watermark location.} Due to the severe tampering within the $\text{\textbf{R}}_{fac}$ after face swapping, avoiding the area during embedding is helpful to preserve the generated watermark. Considering the threatening Case 1, embedding the watermark across the entire $\text{\textbf{R}}_{bac}$ also introduces vulnerabilities especially when attackers immensely crop the background. Given that malicious cropping should not affect the facial attributes within the image, the contour region surrounds the face and encompasses peripheral attributes such as ears and hair (see Fig.~\ref{f3}), which is a beneficial region to embed a watermark resilient to such intentional distortions. Based on the landmark similarity of face contours between the real and swapped images in Fig.~\ref{f2}(b), focusing on the region also offers a feasible solution for constructing watermarks robust to face swapping. Furthermore, localized embedding within the contour ensures visual quality and semantic consistency, as the majority of regions in watermarked images are consistent with the originals. To make the watermark localize the contour region, upon detecting the facial landmarks of $\textbf{X}^{r}$, a facial mask $\textbf{M}_{fac}^{r}$ is generated in Fig.~\ref{f4_new}. Based on the morphological dilation function $\text{MD}(\cdot)$, the contour mask $\text{\textbf{M}}_{c}^{r}$ is derived as $\text{MD}(\textbf{M}_{fac}^{r})-\textbf{M}_{fac}^{r}$, which belongs to the sub-regions of the background and serves as the constraint for watermark embedding. Since some landmark discrepancies between real and swapped images lead to contour mark inconsistency, the iteration of the morphological dilation is randomized during the training stage to simulate the contour mask diversity.

\noindent\textbf{Watermark message.} The determination between real and fake images from previous proactive methods is based on the matching degree between the preset and decoded messages. As non-watermarked images naturally own the mismatching property, it is unreasonable to classify non-watermarked images posted on websites as watermarked fakes. The contour textures—being similar in both real and swapped images—can be selected as one type of message to effectively filter out non-watermarked images. Moreover, considering the identity discrepancy between the real and swapped images, face swapping detection is achieved by comparing the messages involved in identity information. Consequently, the extracted $\text{\textbf{R}}_{c}^{r}$ and $\text{\textbf{R}}_{fac}^{r}$ are used to generate contour and identity features through the pretrained models \cite{lpips,arcface}. Given that the features are represented in high dimensions, embedding them fully into the image proves challenging and brings irrelevant noises. Global average pooling (GAP) and principal component analysis (PCA) \cite{pca} are thus applied to squeeze the spatial and channel dimensions of the extracted features, respectively. To account for inconsistent magnitudes among input images, the squeezed features are scaled to $[-1,1]$. Each element of the scaled identity ${\textbf{F}}_{i}^{r}$ and contour features ${\textbf{F}}_{c}^{r}$$\in$$\mathbb{R}^{1\times C}$ is binarized into ${\textbf{V}}_{i}^{r}$ and ${\textbf{V}}_{c}^{r}$ using a threshold of 0. The hybrid message $\textbf{V}^{r}$ is allocated by concatenating ${\textbf{V}}_{c}^{r}$ and ${\textbf{V}}_{i}^{r}$ as $\textbf{V}^{r}$$=$$[{\textbf{V}}_{c}^{r}$;${\textbf{V}}_{i}^{r}]$. To ensure imperceptible embedding, the magnitude of $\textbf{V}^{r}$ is regulated by a factor $\alpha$ as $\textbf{V}_{\alpha}^{r}$$\in$$\{-\alpha,\alpha\}^{1\times 2C}$.

\noindent\textbf{Embedding process.} The process is performed by integrating the $\textbf{V}_{\alpha}^{r}$ into the original real image. The watermark encoder $\mathcal{E}$ utilizes an alike UNet architecture to generate encoded images. As to the message $\textbf{V}_{\alpha}^{r}$, a combination of multilayer perceptron (MLP) with convolution layers is employed to modulate $\textbf{V}_{\alpha}^{r}$ at different scales, facilitating the integration of the corresponding image features. To ensure the watermark is localized within the $\textbf{R}^{r}_{c}$, the integration at the $k$-th layer is constrained by $\textbf{M}_{c}^{r}$ as follows:
\begin{equation}
\textbf{h}_{k}^{'}=\text{SE}([\text{Conv}_{3\times3}(\text{MLP}(\textbf{V}_{\alpha}^{r}))\odot \text{Rz}(\textbf{M}_{c}^{r});\textbf{h}_{k}]),
\label{e3}
\end{equation}
where SE($\cdot$) represents the squeeze-excitation module \cite{se}. Rz($\cdot$) is the resize function. $\textbf{h}_{k}$ and $\textbf{h}_{k}^{'}$ indicate the original and integrated features at the $k$-th layer. After integrating the $\textbf{V}_{\alpha}^{r}$ into $\textbf{X}^{r}$, the optimization to the $\mathcal{E}$ involves the discrepancies between $\textbf{X}^{r}$ and $\mathcal{E}(\textbf{X}^{r})$, which is expressed as follows:
\begin{equation}
\mathcal{L}_e = l_\text{MSE}(\textbf{X}^{r}, \mathcal{E}(\textbf{X}^{r}))+\lambda_1 l_\text{Adv}(\textbf{X}^{r}, \mathcal{E}(\textbf{X}^{r})),
\label{e4}
\end{equation}
where $\lambda_1$ is hyperparameter. MSE denotes mean square error. $l_\text{Adv}$ indicates the adversarial loss as follows:
\begin{equation}
l_\text{Adv}(\textbf{X}^{r}, \mathcal{E}(\textbf{X}^{r}))\!=\!-\log (Dis(\textbf{X}^{r}))\!-\!\log (1\!-\!Dis(\mathcal{E}(\textbf{X}^{r}))),
\label{e5}
\end{equation}
where $Dis$ is the discriminator from \cite{progan}. After watermark embedding, the watermarked image $\hat{\textbf{X}}^{r}$ is obtained by following:
\begin{equation}
\hat{\textbf{X}}^{r} = \textbf{X}^{r} \odot (1-\textbf{M}_{c}^r) + \mathcal{E}(\textbf{X}^{r}) \odot \textbf{M}_{c}^r.
\label{e6}
\end{equation}
Based on Eq.~\ref{e6}, the watermark involves the designated contour region, which not only ensures the availability of the watermark but also further decreases the visual discrepancy with the original images.
\subsection{Robust Watermark Decoding}
\noindent\textbf{Training stage.} Considering the threatening Case 2, various normal distortions $\mathcal{N}_2(\cdot)$ are introduced to ensure the robust watermark embedded in the $\text{\textbf{R}}_{c}^{r}$. With a large batch size, each iteration will traverse most of the introduced distortions during training, thereby enabling a comprehensive enhancement in watermark resilience. Face swapping techniques are excluded to mitigate the CMark model overfitting to any specific techniques. The categories of normal distortions are described in the implementation details.
\par The $\textbf{M}_{c}^r$ is utilized to isolate irrelevant image regions, thereby preventing interference during decoding. Since the decoding process also requires image analysis, the decoder shares a similar structure with the $\mathcal{E}$ except the message integration. Additionally, to distinguish the non-watermarked images, the optimization of the decoder $\mathcal{D}$ is formulated as follows:
\begin{equation}
 \mathcal{L}_d\!=\!||\mathcal{D}(\textbf{X}^{r}\! \odot\textbf{M}_{c}^r)||_2^2\!+\!\lambda_2 l_\text{MSE}(\textbf{V}_{\alpha}^{r},\mathcal{D}(\mathcal{N}_2(\hat{\textbf{X}}^{r})\odot\textbf{M}_{c}^r)),
 \label{e7}
\end{equation}
where $\lambda_2$ is the hyperparameter. The first term represents the decoder output for the non-watermarked real image $\textbf{X}^{r}$, which is regularized to zero to distinguish the message allocated to the corresponding watermarked image as each element of the message towards 1 or -1. To make the watermark imperceptible and robust, the $\mathcal{E}$ and $\mathcal{D}$ are trained end-to-end via $\mathcal{L}_e + \mathcal{L}_d$.
\begin{figure}
\centering
\includegraphics[width=\linewidth]{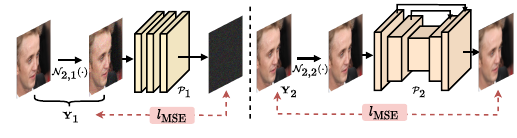}
\caption{Illustration of the proposed denoiser modules, $\mathcal{P}_1$ and $\mathcal{P}_2$, during the training phase.}
\label{f5_new}
\end{figure}
\par In practical decoding process, the contour mask employed comes from the manipulated $\hat{\textbf{X}}^{r}$. The normal distortions cause the contour mask extracted from the noised images deviates from the clean counterparts, thereby impacting the decoding correctness. To decrease the mask biases, a proactive denoiser $\mathcal{P}$ is introduced to restore normal distorted images. The initial module $\mathcal{P}_1$ in Fig. \ref{f5_new} estimates randomly distributed noise, such as Gaussian Noise, while $\mathcal{P}_2$ recovers details lost due to blurring or compression. The supervision $\textbf{Y}_1$ $\in$ $\mathbb{R}^{H\times W\times 3}$ corresponds to the residual mapping $\textbf{X}^r - \mathcal{N}_{2,1}(\textbf{X}^r)$, where $\mathcal{N}_{2,1}(\cdot)$ denotes random noise. The supervision $\textbf{Y}_2$ equals $\textbf{X}^r$ for denoiser module $\mathcal{P}_2$ and $\mathcal{N}_{2,2}(\cdot)$ involves blur or compression. The optimization of the $\mathcal{P}$ is expressed as follows:
\begin{equation}
 \mathcal{L}_{p}\!=\!l_\text{MSE}(\textbf{Y}_1,\!\mathcal{P}_1(\mathcal{N}_{2,1}(\textbf{X}^r)))\!+\!l_\text{MSE}(\textbf{Y}_2,\!\mathcal{P}_2(\mathcal{N}_{2,2}(\textbf{X}^r))).
 \label{e9}
\end{equation}
\textbf{Testing stage.} After generating the watermarked image $\hat{\textbf{X}}^r$, the denoised image $\hat{\textbf{X}}_d$ is computed by:
\begin{equation}
 \hat{\textbf{X}}_d = \mathcal{P}_2(\mathcal{P}_1(\mathcal{N}(\hat{\textbf{X}}^r)) + \mathcal{N}(\hat{\textbf{X}}^r)),
 \label{e10}
\end{equation}
where $\mathcal{N}(\cdot)$ encompasses face swapping $\mathcal{N}_1(\cdot)$ and normal distortions $\mathcal{N}_2(\cdot)$. Based on the image $\mathcal{N}(\hat{\textbf{X}}^r)$ and the extracted mask $\hat{\textbf{M}}_{c}$ from the denoised image $\hat{\textbf{X}}_d$, the decoded message $\hat{\textbf{V}}$$\in$$\{-1, 1\}^{1\times 2C}$ is obtained by binarizing the result $\mathcal{D}(\mathcal{N}(\hat{\textbf{X}}^r)\odot\hat{\textbf{M}}_{c})$.

\subsection{Encrypted Message Verification}
To achieve self-verification and thus avoid the storage of massive preset messages, we verify the decoded message $\hat{\textbf{V}}$ with the reference counterpart $\bar{\textbf{V}}$$\in$$\{-1, 1\}^{1\times 2C}$. Since the normal distortions interfere with verification, $\bar{\textbf{V}}$ comes from the denoised image $\hat{\textbf{X}}_d$ and is extracted based on the same pretrained models \cite{arcface,lpips}. Considering the threatening Case 3, these messages should be encrypted to safely transmit to the platform for verification. To prevent the leakage of both the ${\hat{\textbf{V}}}$ and $\bar{\textbf{V}}$ during communication, two distinct public keys are employed for encryption. This scheme ensures that attackers can not deduce the determined result upon intercepting these encrypted messages, as the encrypted $\hat{\textbf{V}}$ and $\bar{\textbf{V}}$ remain different regardless of whether their messages are consistent.
\begin{algorithm}[t]
\caption{Encrypted Message Verification}
\begin{algorithmic}[1]
\REQUIRE 
$\textbf{E}_{\hat{v}}$: Encrypted decoded message; 
$\textbf{E}_{\bar{v}}$: Encrypted reference message; 
$t_1$, $t_2$: Detection thresholds; 
$d_1$, $d_2$: Private keys;
$p_1$, $p_2$, $q_1$, $q_2$: Prime numbers
\ENSURE 
$O$: Predicted output of the manipulated image
\STATE $O\leftarrow\text{None}$
\STATE $\hat{\textbf{V}}\!=\!\text{RSA}\_\text{De}(\textbf{E}_{\hat{v}},\!q_1,\!p_1,\!d_1)$,\! $\bar{\textbf{V}}\!=\!\text{RSA}\_\text{De}(\textbf{E}_{\bar{v}},\!q_2,\!p_2,\!d_2)\!$
\STATE ${\hat{\textbf{V}}}_{i} \leftarrow {\hat{\textbf{V}}}[:,0:C-1]$, $\hat{\textbf{V}}_{c} \leftarrow {\hat{\textbf{V}}}[:,C:2C-1]$
\STATE $\bar{\textbf{V}}_{i} \leftarrow \bar{\textbf{V}}[:,0:C-1]$,  $\bar{\textbf{V}}_{c} \leftarrow \bar{\textbf{V}}[:,C:2C-1]$
\IF[Watermark Judgment]{BER(${\hat{\textbf{V}}}_{c}$, $\bar{\textbf{V}}_{c}$)$\leq$$t_1$}
    \IF[Authenticity Judgment]{BER(${\hat{\textbf{V}}}_{i}$, $\bar{\textbf{V}}_{i}$)$\leq$$t_2$}
        \STATE $O\leftarrow\text{Real}$
    \ELSE
        \STATE $O\leftarrow\text{Fake}$
    \ENDIF  
\ENDIF   
\RETURN $O$
\end{algorithmic}
\label{a1}
\end{algorithm}
\par Given the potential for key exposure on the client sides, we adopt an asymmetric strategy by assigning public keys to the clients for encryption and corresponding private keys to the platform for decryption. Based on the Rivest-Shamir-Adleman (RSA) algorithm \cite{rsa}, the message encryption, decryption, and key generation are defined as RSA$\_$En$($$\cdot$$)$, RSA$\_$De$($$\cdot$$)$, and RSA$\_$KeyGen$($$\cdot$$)$. Let $\phi=(p-1)\cdot(q-1)$, where $p$ and $q$ are different prime numbers, a pair of public key $e$ and private key $d$ are generated by $\text{RSA}\_\text{KeyGen}(\phi)$. The $\mathcal{D}$ is assigned a key $e_1$, while the face identity and contour texture extractors share another key $e_2$. The encrypted messages $\textbf{E}_{\hat{v}}$ and $\textbf{E}_{\bar{v}}$ \textit{w.r.t.} $\hat{\textbf{V}}$ and $\bar{\textbf{V}}$ are generated as:
\begin{equation}
\textbf{E}_{\hat{v}}\!=\!\text{RSA}\_\text{En}(\hat{\textbf{V}},q_1,p_1,e_1);\textbf{E}_{\bar{v}}\!=\! \text{RSA}\_\text{En}(\bar{\textbf{V}},q_2,p_2,e_2).
\label{e13}
\end{equation}
\par The image can be progressively determined based on the proposed hybrid message. In Algorithm \ref{a1}, after decryption with private keys, $\hat{\textbf{V}}_{c}$ is extracted from the first $C$ elements of $\hat{\textbf{V}}$ and verified against the $\bar{\textbf{V}}_{c}$ to determine whether the watermark is embedded. If the image is confirmed to be watermarked, the identity message ${\hat{\textbf{V}}}_{i}$, derived from the rest elements of ${\hat{\textbf{V}}}$, is verified against ${\bar{\textbf{V}}}_{i}$ to access the image authenticity. Bit error rate (BER) is formulated as follows:
\begin{equation}
\text{BER}(\hat{\textbf{V}},\bar{\textbf{V}}) = \frac{\sum_{i=0}^{l-1} \mathbbm{1}({\hat{\textbf{V}}}[i]=\bar{\textbf{V}}[i])}{l} \times 100\%,
\label{e12}
\end{equation}
where $l$ denotes message length. $\mathbbm{1}(\hat{\textbf{V}}[i]$$=$$\bar{\textbf{V}}[i])$ is the indicator function, which equals 1 if $\hat{\textbf{V}}[i]$=$\bar{\textbf{V}}[i]$, and 0 otherwise. More details of incorporating the RSA algorithm are presented in Appendix~\ref{secA1}.

% === III. Schottky-Diode Class-C Rectifier =======================================
% =================================================================================

\begin{table*}[t]
    \centering
    \renewcommand{\arraystretch}{1.2}
    \renewcommand\tabcolsep{1.4pt}
    \footnotesize
    \caption{Comparisons of detection performance on CelebA-HQ \cite{celebhq}. The best and secondary results are denoted as \textbf{bold} and \underline{underline}. $\dag$ denotes the model is tested on 256$\times$256. Since there are two-step determinations, we regard the watermarked image is wrong during the statistics of ACC and F1-Score when BER$_{c}^{re}$$>$$t_1$. `Pas'. and `Pro.' denote passive and proactive detectors. `SV.' means self-verification.}
    \scalebox{0.75}{
    \begin{tabular}{c|c|c|cc|cc|cc|cc|cc|cc|cc|cc|cc}
    \toprule[1.2pt]
        \multirow{2}*{Method}& \multirow{2}*{Type} & \multirow{2}*{SV.}  & \multicolumn{2}{c}{FaceSwap} & \multicolumn{2}{c}{MobileFS}& \multicolumn{2}{c}{FaceShifter }& \multicolumn{2}{c}{SimSwap} & \multicolumn{2}{c}{MegaFS}& \multicolumn{2}{c}{E4S}& \multicolumn{2}{c}{DiffSwap}& \multicolumn{2}{c|}{DiffFace} & \multicolumn{2}{c}{\textbf{Avg.}$\uparrow$}\\
        \cline{4-21}
        &&
        & F1 & {ACC}& F1 & {ACC} & F1 & {ACC} & F1 & {ACC} & F1 & {ACC} & F1 & {ACC} & F1 & {ACC}& F1 & {ACC} & F1 & ACC\\
        \hline
        {Xception}\cite{ff++}& Pas.&-& 75.56 & 71.89& 71.30 & 67.91 & 57.07 & 56.77 & 58.43 & 59.56 & 80.06 & 76.70 & 66.88 & 61.19 & 74.83 & 71.06 &64.89 & 62.60&  68.63 & 66.38 \\
        {F3Net} \cite{f3}& Pas.&-& 81.03 & \underline{91.91}& 67.64 & 71.96 & 57.11 & 66.33 & 62.59 & 65.60 & 54.68 & 56.16 & 80.17 & 81.02 & 64.33 & 69.85 & 51.36 & 54.23 & 64.86 & 68.38\\
        {RECCE} \cite{recce}& Pas.&-& 74.69 & 66.21& 74.70 & 66.43 & 74.83 & 66.40 & 61.16 & 67.39 & 74.87 & 66.43 & 74.63 & 66.18 & 74.86 & 66.42& 71.05 & 62.50 & 72.62 & 65.99 \\
        {SBI} \cite{sbi}& Pas.&-& 58.44 & 62.08 & 82.66 & 82.18 & 67.67 & 62.90 & 60.98 & 62.01 & 71.13 & 65.98 & 69.26 & 64.30 & 77.63 & 72.24& 67.34 & 62.62 & 69.39 & 66.79 \\
        {CADDM} \cite{caddm}& Pas.&-& 84.56 & 82.44& 84.23 & 82.01 & 80.31 & 78.27 & 70.57 & 69.96 & 80.45 & 78.40 & 80.27 & 78.23 & \underline{84.40} & \underline{82.18} & 80.56 & 78.50  & 80.67 & 78.75 \\
        {FPG} \cite{fpg}& Pas.&-& 70.66 & 71.11 & 83.62 & 81.93 & 73.37 & 73.00 & 71.81 & 71.77 & 82.75 & 81.11 & 79.28 & 77.97 & 77.52 & 71.71 & 77.70 & 76.59 & 77.09 & 75.65 \\
        {ProDet} \cite{prodet}& Pas.&-& 69.28 &64.69 &67.29&60.83& 74.48& 68.89& 68.33& 62.18& 74.64 &68.05 &70.21& 64.42 &68.60 &66.08 &68.42 &65.91 &70.16 &65.16\\
        ODDN \cite{oddn}& Pas.&-& 74.95 &69.97 &70.86 &62.94 &73.22 &67.37 &73.75 &69.51 &68.81 &60.84 &68.19 &61.56 &68.66 &62.52 &68.53 &62.24 &70.87 &64.62\\
        \hline
        SepMark$^\dag$ \cite{sepmark}& Pro.&$\checkmark$& 70.02 & 57.18 & 66.89  & 50.50 & 71.32 & 59.80 & \textbf{100} & \textbf{100} & 70.29 & 57.74 & 70.08 & 57.31 & 72.33 & 61.84 & 77.91 & 71.64 & 74.86 & 64.50 \\
        IDMark$^\dag$ \cite{proactive}& Pro.&$\times$& 86.77 &84.76 & \textbf{97.56}	&\textbf{97.32}&	\underline{94.78}&	\underline{94.05}&	91.06&	90.68&	\underline{96.12}	&\underline{95.96} &\textbf{97.43} &	\textbf{97.45} &	71.57 & 62.75 &93.19&	91.82	&91.06 & 89.35  \\
        \hline
        {CMark} (Ours)& Pro.&$\checkmark$&\underline{91.08}  & {90.37} & {92.99} & {92.58} & {90.66} & {89.86} & {92.44} & {91.96} & {92.59} & {92.13} & {93.18} & {92.80} & {83.44} & {81.28} & \underline{92.61} & \underline{92.15} & \underline{91.12} & \underline{90.39}\\
        {CMark$^\dag$} (Ours)& Pro.&$\checkmark$& \textbf{94.27} & \textbf{93.96} & \underline{97.01} & \underline{96.94} & \textbf{96.62} & \textbf{96.52} & \underline{95.54} & \underline{95.37} & {\textbf{96.57}} & {\textbf{96.46}} & \underline{96.35} & \underline{96.24} & \textbf{85.85} & \textbf{83.62} & \textbf{93.95} & \textbf{93.60} & \textbf{94.52} & \textbf{94.09} \\
     \bottomrule[1.2pt]
    \end{tabular}}
    \label{t1}
\end{table*}

\section{Experiments}
\label{sec:5}
\subsection{Experimental Setup}
\noindent\textbf{Implementation details.}
All images are resized to 128 and 256, respectively. The Adam optimizer \cite{adam} is used for training 100 epochs with $\beta_1$=0.5 and $\beta_2$=0.999. The learning rate is set to 2$e^{-4}$ with the batch size equal to 96. We assign $C$=16 for contour texture and face identity features. There is thus a 32-bit preset message with $\alpha$=0.1 for each image. $\lambda_{1}$ and $\lambda_{2}$ are set as 0.1 and 30. During verification, $t_1$ and $t_2$ are both set as $\frac{3}{16}\times100\%$ as default. The iteration of the morphological dilation is randomly selected within [2, 4] during training. Considering the complex real-world scenarios, we introduce Gaussian Noise, JPEG, Gaussian Blur, Edge Crop, SaltPepper Noise, Median Blur, Resize, Dropout, Brightness, Contrast, Hue, and Saturation distortions during training. All experiments are conducted on four RTX 3090 GPUs.

\noindent\textbf{Dataset.} The experiments are mainly conducted on CelebA-HQ \cite{celebhq} for training and testing. To demonstrate the transferability of our approach, we further perform a cross-dataset evaluation on the FFHQ dataset \cite{stylegan}.

\noindent\textbf{Evaluation metrics.} There are two types of BER. One is the comparison between the decoded message and preset message (\textit{i.e.}, BER($\hat{\textbf{V}}$,$\textbf{V}^r$)). The other is the comparison between the decoded message and reference message (\textit{i.e.}, BER($\hat{\textbf{V}}$,$\bar{\textbf{V}}$)). We briefly denote them as BER$^{pr}$ and BER$^{re}$. Face swapping detection involves accuracy, F1-Score, and BER$^{re}$, while watermark robustness is evaluated using BER$^{pr}$. We apply PSNR, SSIM, and LPIPS \cite{lpips} to assess the quality of the watermarked images. CMark is compared with benchmarks that release code for fairness.
\begin{figure}
\centering
\includegraphics[width=\linewidth]{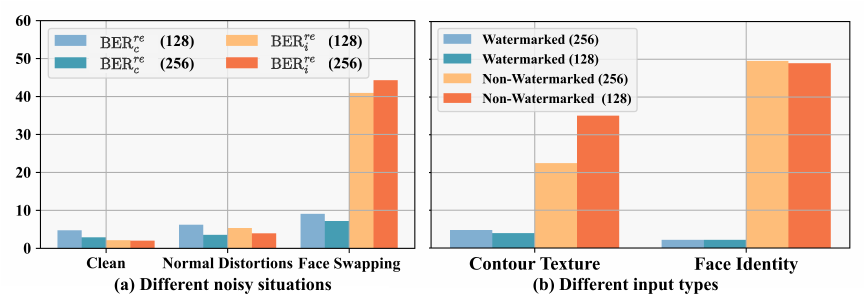}
\caption{\textbf{(a)}. Average BER$_{c}^{re}$ and BER$_{i}^{re}$ on different normal image distortions and face swapping techniques. `Clean' means no image distortions. \textbf{(b)}. BER$^{re}$ on watermarked and non-watermarked images.}
\label{f4}
\end{figure}
\subsection{Effectiveness Evaluation} 
\noindent\textbf{Face swapping detection.}
\begin{table*}[t]
    \centering
    \footnotesize
    \renewcommand{\arraystretch}{1.2}
    \renewcommand\tabcolsep{1.0pt}
    \caption{Comparisons of BER$^{pr}$ under different face swapping techniques. `CQ' and `FQ' indicate CelebA-HQ \cite{celebhq} and FFHQ \cite{stylegan} datasets.  $\dag$ and $\ddag$ denote the model is evaluated on 256$\times$256 and 512$\times$512. The best and secondary results are denoted as \textbf{bold} and \underline{underline}.}
    \scalebox{0.9}{
    \begin{tabular}{c|cc|cc|cc|cc|cc|cc|cc|cc|cc}
    \toprule[1.2pt]
    \multirow{2}*{Method}& \multicolumn{2}{c}{FaceSwap} & \multicolumn{2}{c}{MobileFS}& \multicolumn{2}{c}{FaceShifter}& \multicolumn{2}{c}{SimSwap} & \multicolumn{2}{c}{MegaFS}& \multicolumn{2}{c}{E4S}& \multicolumn{2}{c}{DiffSwap}& \multicolumn{2}{c|}{DiffFace} & \multicolumn{2}{c}{\textbf{Avg.}$\downarrow$}\\
    \cline{2-19} & CQ & FQ & CQ & FQ & CQ & FQ & CQ & FQ & CQ & FQ & CQ & FQ & CQ & FQ & CQ & FQ & CQ & FQ\\
    \hline
    HIDDEN \cite{zhu2018hidden}  &16.87 &18.89 &12.48&12.08 &16.24&15.31&27.89&29.49	&11.59 &11.97	&11.49	&11.69 &24.78	& 26.53 &11.79 & 11.38 & 16.64 & 17.17\\
    PIMOG \cite{pimog}  &8.45&11.28	&0.40 &\underline{0.31} &1.23	&\textbf{1.01} &8.68&\textbf{2.99}	&15.55	& 17.33&13.07	&15.37 &17.83	& 8.25 & 19.68&18.71	&10.61 & 9.41\\
    ARWGAN \cite{arwgan}  &2.41	& 4.65 &1.70&0.87	&2.61&1.49	&23.67&25.56	&0.96	& 0.92&\underline{1.06}	& 0.96 &32.20&18.02	&1.16	& \textbf{0.04}&8.22 &6.56 \\
    SepMark \cite{sepmark}  &\textbf{0.43}& \underline{2.87}	&\textbf{0.05}	& \textbf{0.04} &\textbf{0.53}	& \underline{1.37} &\underline{4.89}	&5.97&\underline{0.36}&\underline{0.84}	&0.11	&\underline{0.37} &\textbf{0.10}&\underline{0.66}	&\textbf{0.17}&\underline{0.84} &\underline{0.83}&\underline{1.62}\\
    CMark (Ours)  &\underline{1.24}	& \textbf{0.32} &\underline{0.21}& 0.51&\underline{0.85}&2.83&\textbf{2.60}&\underline{5.11}	&\textbf{0.29}	& \textbf{0.75} & \textbf{0.11} & \textbf{0.34}	&\underline{0.14}&\textbf{0.45}	&\underline{0.23}&2.27	&\textbf{0.71}&\textbf{1.57}\\
    \hline
    MBRS$^\dag$ \cite{mbrs}  &13.10	&14.62 &\textbf{0.98}	&\textbf{0.94}&\underline{1.23}	&\underline{0.69}&46.73&47.26&13.96&15.01	&13.97&14.96	&41.41&\underline{9.11} &12.36&12.78	&17.96 &14.42\\
    FaceSigns$^\dag$ \cite{facesigns}  &50.16&50.09 &50.47&49.97	&48.91	&49.92&50.24&49.78	&49.50&49.01	&50.27&49.41 &46.93&45.51 &45.37 &44.45	&48.98&48.51\\
    SepMark$^\dag$ \cite{sepmark} & \underline{8.70} & \underline{2.15}	& 11.21 & 7.69	&\textbf{0.29} &\textbf{0.62}	&\textbf{7.91}&\textbf{7.69}	&{4.64} &{5.32} &{4.79}&{5.59}	&\underline{17.33}	&18.36&\textbf{0.36}	&\underline{2.61}&\underline{6.90}&\underline{6.25}\\
    EditGuard$^\ddag$ \cite{editguard} & 84.55 &87.27 &70.32 &71.70 &74.54 &74.41 &48.85 &49.59 &\underline{2.75} &\textbf{2.26}  &\textbf{0.21} &\textbf{0.26} &75.87 &77.89 &59.46 &60.93 &52.06 &53.03\\
    CMark$^\dag$ (Ours)  &\textbf{1.42}&\textbf{0.59}	&\underline{2.29}&\underline{3.01}	&1.71&3.14	&\underline{10.16}&\underline{13.53}	&\textbf{1.66}&\underline{2.45}	&\underline{2.03}&\underline{2.94}	&\textbf{1.54}&\textbf{1.63} &\underline{0.65}&\textbf{1.42}	&\textbf{2.68}& \textbf{3.58}\\
    \bottomrule[1.2pt]
    \end{tabular}}
    \label{t4}
\end{table*}
A comprehensive evaluation in comparison with state-of-the-art passive and proactive detectors is shown in Tab.~\ref{t1}. After embedding only 32-bit messages into the images, the average accuracy and F1-Score across 8 face swapping techniques exceed the second-best passive detector over 11.64\% and 10.45\%, respectively. More importantly, the generalization of our approach is better than other proactive detectors, and this is further demonstrated at 256$\times$256 as the embedding space is expanded. Since SepMark \cite{sepmark} is trained with SimSwap \cite{simswap}, it attains excellent performance in this technique but loses generalization. Although IDMark \cite{proactive} demonstrates comparable performance, its detection performance to DiffSwap \cite{diffswap} is severely degraded. The lacking of self-verification also leads to massive storage of preset messages. Additionally, it exhibits a poor visual quality for watermarked images (see Tab.~\ref{t3}). These findings underscore the effectiveness of CMark which simultaneously achieves generalized detection and invisible watermark generation. Fig.~\ref{f4}(a) illustrates the BER$^{re}\!$ under different situations. BER$^{re}\!$ consists of the face identity error rate BER$^{re}_{i}$ and the contour texture error rate BER$^{re}_{c}\!$. Apparently, the low BER$_{c}^{re}\!$ across different situations indicates the similarity of contour texture between real and swapped images. After face swapping, due to the biases in different techniques, the synthesized backgrounds may deviate from the originals and slightly fluctuate the BER$_{c}^{re}\!$. In contrast, there is a proliferation in BER$_{i}^{re}\!$, enabling us to set a straightforward threshold for detection rather than building complex classification heads trained in different face swapping techniques.

\begin{table}[t]
    \centering
    \renewcommand{\arraystretch}{1.2}
    \renewcommand\tabcolsep{1.2pt}
    \footnotesize
    \caption{Comparisons of BER$^{pr}$ under normal distortions on CelebA-HQ \cite{celebhq}. $\dag$ and $\ddag$ denote the model is evaluated on 256$\times$256 and 512$\times$512.}
    \scalebox{0.85}{
    \begin{tabular}{c|cccc|c}
    \toprule[1.2pt]
    Method & GNoise & GBlur & ECrop & JPEG & \textbf{Avg.} $\downarrow$ \\
    \hline
    HIDDEN \cite{zhu2018hidden} &48.64	&26.96	&12.21	&32.16	&29.99\\
    PIMOG \cite{pimog}  &12.73	&0.12	&1.74	&19.56	&8.53\\
    ARWGAN \cite{arwgan}  &46.55	&14.78	&2.37	&42.58	&26.57\\
    SepMark \cite{sepmark}  &\underline{0.82}&\textbf{0.01}&\textbf{0.01}&\textbf{0.24} &\underline{0.27}\\
     CMark (Ours)  &\textbf{0.47}&\underline{0.03}&\underline{0.03}&\underline{0.46} &\textbf{0.25}\\
    \hline
    MBRS$^\dag$ \cite{mbrs}  &41.68&27.95&19.33&0.31 &22.32\\
    FaceSigns$^\dag$ \cite{facesigns}  &0.86&	0.17&	\underline{0.27}&	0.85&	\underline{0.54}\\
    SepMark$^\dag$ \cite{sepmark}  &\textbf{0.06}&\textbf{0.01}&18.13& \textbf{0.01} &4.55\\
    EditGuard$^\ddag$ \cite{editguard} &15.81&98.31&3.94&23.62&35.42\\
    CMark$^\dag$(Ours)  &\underline{0.31}&\underline{0.12}&\textbf{0.12}  & \underline{0.73} &\textbf{0.32}\\
    \bottomrule[1.2pt]
    \end{tabular}}
    \label{t5}
\end{table}

\begin{figure}
\centering
\includegraphics[width=\linewidth]{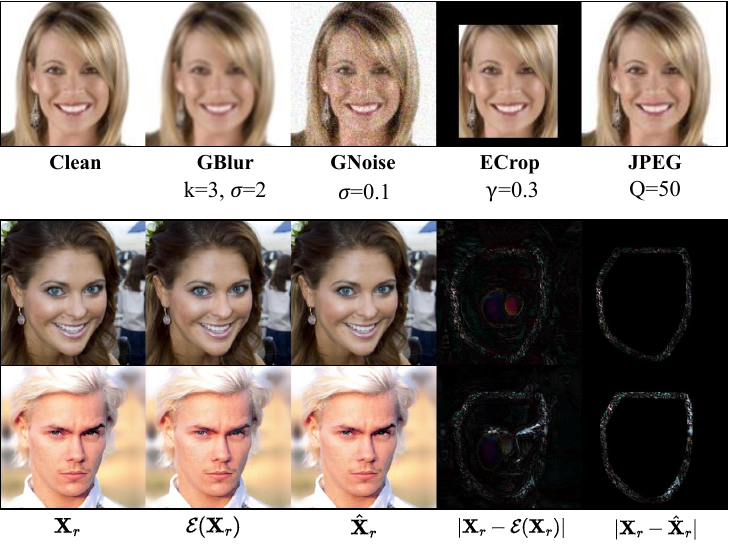}
\caption{The \textbf{first row} represents the watermarked image under normal distortions. The \textbf{last two rows} are the visual comparisons among the original real, encoded, and watermarked images. `GNoise', `GBlur', and `ECrop' denote Gaussian noise, Gaussian blur, and edge crop. k, $\sigma$, $\gamma$, and Q denote the kernel size, standard deviation, cropped ratio, and compression quality.}
\label{f5}
\end{figure}

\noindent\textbf{Watermark robustness.}
Considering the watermark robustness is necessary to ensure accurate face swapping detection, the results on different datasets are summarized in Tab. \ref{t4}. Apparently, our approach achieves the lowest average BER$^{pr}$ than other methods, confirming the robustness across various unknown face swapping techniques. Robustness evaluation also involves four representative image distortions that are presented in the first row of Fig.~\ref{f5}. While some methods can achieve better PSNR, the results in Tab. \ref{t5} indicate that they are ineffective against normal distortions, which hinders the practicality. The CMark model attains the lowest average BER$^{pr}$, demonstrating superior robustness across normal distortions. These results underscore a satisfactory trade-off between visual quality and watermark robustness. The BER$^{pr}$ results under the more diverse normal distortions are listed in Appendix~\ref{secA3}.

\noindent\textbf{Visual quality.}
 As illustrated in the last two rows of Fig.~\ref{f5}, the encoded images exhibit imperceptible discrepancies, making them indistinguishable from the human eyes. Leveraging Eq.~\ref{e6} further decreases the visual differences. Tab.~\ref{t3} reports the watermarked images generated from the CMark model achieves superior SSIM along with competitive PSNR and LPIPS. The degraded quality is attributed to the limited embedding location, which restricts the integration of the preset messages. Nevertheless, the watermarked images are lossless within the facial regions, effectively preserving the identity of the corresponding original images for downstream face analysis \cite{expression1,face_attribute}. More qualitative results of the watermarked images on different datasets are illustrated in Appendix~\ref{secA2}.

\begin{table}[t]
    \centering
    \renewcommand{\arraystretch}{1.2}
    \renewcommand\tabcolsep{1.1pt}
    \footnotesize
    \caption{Comparisons of visual quality on CelebA-HQ \cite{celebhq}.}
    \scalebox{0.78}{
    \begin{tabular}{c|c|c|ccc}
    \toprule[1.2pt]
    Method & Resolution &Length &PSNR$\uparrow$ &LPIPS $\downarrow$ &SSIM $\uparrow$\\
    \hline
    HIDDEN \cite{zhu2018hidden} &128$\times$128&30 bit&33.44&0.0120& 0.900\\
    PIMOG \cite{pimog}  &128$\times$128&30 bit&37.73&0.0086&0.947 \\
    SepMark \cite{sepmark}  &128$\times$128&30 bit&38.51&\textbf{0.0028}&0.959  \\
    ARWGAN \cite{arwgan}  &128$\times$128&30 bit&\textbf{40.23}&0.0063&\underline{0.969}\\
    EditGuard \cite{editguard}  &128$\times$128&30 bit&36.93&-&0.944 \\
    CMark (Ours)  &128$\times$128&32 bit&\underline{39.36}&\underline{0.0062}& \textbf{0.972}\\
    \hline
    MBRS \cite{mbrs}  &256$\times$256&256 bit&\textbf{44.03}&\textbf{0.0045}&\underline{0.972} \\
    FaceSigns \cite{facesigns}  &256$\times$256&128 bit&32.33&0.0318&0.921 \\
    IDMark \cite{proactive}  &256$\times$256&512 bit&28.26&0.0692&0.741 \\
    SepMark \cite{sepmark}  &256$\times$256&128 bit&38.56&0.0080&0.933 \\
    CMark (Ours)  &256$\times$256&32 bit& \underline{41.86} & \underline{0.0068}& \textbf{0.983}\\
    \bottomrule[1.2pt]
    \end{tabular}}
    \label{t3}
\end{table}

\subsection{Evaluation under Various Scenarios}
\noindent\textbf{Non-watermarked images.} With the first term specified in Eq.~\ref{e7}, the decoder outputs for non-watermarked images are regularized to 0 and deviate from the magnitude of each element in the preset messages, which promotes distinguishing the non-watermarked images. Fig.~\ref{f4}(b) shows the discrepancies of BER$^{re}$ when inputting the decoder with the watermarked and non-watermarked images. Noticeably, non-watermarked images yield higher BER$^{re}$ than watermarked ones. %Based on $\!t_1$, the detection accuracy of watermarked and non-watermarked images is 93.51\% and 84.62\%, which enables the filtering of non-watermarked images.

\noindent\textbf{Cross-Dataset evaluation.}
\begin{table}[t]
    \centering
    \footnotesize
    \renewcommand{\arraystretch}{1.0}
    \renewcommand\tabcolsep{7.5pt}
    \caption{Evaluation of cross-dataset performance with different image resolutions on FFHQ \cite{stylegan} dataset.}
    \begin{tabular}{c|c|cc}
    \toprule[1.2pt]
    \multirow{2}*{Metric} & \multirow{2}*{Manipulations} & \multicolumn{2}{c}{FFHQ \cite{stylegan}}\\
    \cline{3-4}
    &&128$\times$128&256$\times$256\\
    \hline
    \multirow{4}*{BER$^{pr}$$\downarrow$}&GNoise&0.46&0.47\\
    &GBlur&0.06&0.17\\
    &ECrop&0.09&0.19\\
    &JPEG&0.60&0.82\\
    \cline{2-4}
    &\textbf{Avg.}&0.30&0.41\\
    \hline
    \multirow{8}*{ACC$\uparrow$}&FaceSwap &90.89&93.46\\
    &MobileFS &91.58&96.58\\
    &FaceShifter &90.29&95.92\\
    &SimSwap  &91.33&95.43\\
    &MegaFS  &92.94&96.32\\
    &E4S &93.37&97.14\\
    &DiffSwap &80.34&82.76\\
    &DiffFace &90.96&93.50\\
    \cline{2-4}
    &\textbf{Avg.}&90.21&93.89\\
    \bottomrule[1.2pt]
    \end{tabular}
    \label{t7}
\end{table}
Besides the Tab.~\ref{t4}, to assess the transferability of our approach, the CMark model is further evaluated on the FFHQ \cite{stylegan} dataset with different normal distortions and face swapping techniques. Given that contour texture and face identity information are derived from generalized models \cite{lpips,arcface}, the extracted squeezed features exhibit slight deviations. Furthermore, the watermark encoding and decoding processes are confined to the contour region within the image, thereby reducing the impact of variation across different datasets. Consequently, the results from Tab.~\ref{t7} demonstrate similar performance to those observed on the CelebA-HQ \cite{celebhq} dataset at the 128 and 256 resolutions, respectively. As for visual quality, it can be seen from Tab.~\ref{t11} that the quantitative results are similar to those of these two datasets, which demonstrates the transferability of our approach. Due to the embedding strategy in the CMark model, there are high identity similarities across different datasets, ensuring the consistency of face analysis between the watermarked images and the corresponding originals. The qualitative results on FFHQ \cite{stylegan} are presented in Fig.~\ref{s_f3} of Appendix~\ref{secA2}. 

\begin{table}[t]
    \centering
    \footnotesize
    \renewcommand{\arraystretch}{1.2}
    \renewcommand\tabcolsep{1pt}
    \caption{Comparison of visual quality under different resolutions on CelebA-HQ \cite{celebhq} and FFHQ \cite{stylegan} datasets. `IS' denotes identity similarity compared with the corresponding original images.}
    \label{t11}
    \scalebox{0.82}{
    \begin{tabular}{c|cccc|cccc}
    \toprule[1.2pt]
     \multirow{2}*{Resolution}&\multicolumn{4}{c|}{CelebA-HQ \cite{celebhq}}&\multicolumn{4}{c}{FFHQ \cite{stylegan}}\\
     \cline{2-9}
     &PSNR $\uparrow$& LPIPS $\downarrow$ & SSIM $\uparrow$ & IS $\uparrow$& PSNR $\uparrow$& LPIPS $\downarrow$ & SSIM$\uparrow$ & IS $\uparrow$\\
     \hline
     128$\times$128& 39.36 & 0.0062 & 0.972 &0.993 &37.53& 0.0048&0.962&0.986\\
     256$\times$256 & 41.86 & 0.0068 & 0.983 &0.996 &41.19& 0.0074&0.977&0.991\\
    \bottomrule[1.2pt]
    \end{tabular}}
\end{table}

\begin{table}[t]
    \centering
    \footnotesize
    \renewcommand{\arraystretch}{1.2}
    \renewcommand\tabcolsep{3.5pt}
    \caption{Evaluation of detection robustness on CelebA-HQ dataset \cite{celebhq}.}
    \begin{tabular}{c|c|cccc|c}
    \toprule[1.2pt]
    Swapping&\multirow{2}*{None}&\multicolumn{5}{c}{Normal Distortions (ACC $\uparrow$)}\\
    \cline{3-7}
    Techniques&&GNoise&GBlur&ECrop&JPEG&\textbf{Avg.}\\
    \hline
    FaceSwap& 90.37& 90.21	&88.23	&88.58&	88.78&88.95\\
    MobileFS & 92.58& 92.09	&90.53	&91.52	&90.69&91.21\\
    FaceShifter&89.86&88.49	&81.52	&87.68	&82.38&85.02\\
    SimSwap &91.96 & 85.02 & 87.48 & 85.80 & 87.51&86.45\\
    MegaFS &92.13&92.09	&90.62	&91.76	&90.22&91.17\\
    E4S &92.80&92.32  &91.34	&91.79	&91.25&91.68\\
    DiffSwap &81.28&81.04	&80.85	&80.59	&80.16	&80.66 \\
    DiffFace &92.15&91.45	&90.89	&90.38	&90.80	&90.88\\
    \bottomrule[1.2pt]
    \end{tabular}
    \label{t6}
\end{table}

\begin{table}[t]
    \centering
    \footnotesize
    \renewcommand{\arraystretch}{1.2}
    \renewcommand\tabcolsep{1pt}
    \caption{Comparisons of BER$^{pr}$ under SimSwap \cite{simswap} with different image normal distortions on CelebA-HQ \cite{celebhq}. $\dag$ and $\ddag$ denote the model is tested on 256$\times$256 and 512$\times$512.}
    \scalebox{0.86}{
    \begin{tabular}{c|cccc|c}
    \toprule[1.2pt]
    \multirow{2}*{Method}&\multicolumn{5}{c}{SimSwap (BER$^{pr}$ $\downarrow$)}\\
    \cline{2-6}
    &GNoise&GBlur&ECrop&JPEG&\textbf{Avg.}\\
    \hline
    HIDDEN \cite{zhu2018hidden} &28.76 &47.47	&39.08	&46.89	&40.55\\
    PIMOG \cite{pimog}  &21.88 & 46.56  &32.96  &47.92  &37.33\\
    ARWGAN\cite{arwgan}  &\textbf{2.34}  &24.30    &\underline{6.64}  &29.36  &15.66\\
    SepMark \cite{sepmark}  &2.76  &\underline{14.83}  &21.82   &\textbf{4.13}  &{\underline{10.89}}\\
    CMark (Ours) &\underline{2.48}   &\textbf{9.77} &  
    \textbf{4.49}   &\underline{7.54}   &\textbf{6.07} \\
    \hline
    MBRS$^\dag$ \cite{mbrs}  &49.38&45.71&49.12&49.05 &48.31\\
    FaceSigns$^\dag$ \cite{facesigns}  &49.76&	50.01&	49.98&	49.83&49.89\\
    SepMark$^\dag$ \cite{sepmark}  &\underline{32.26} & \underline{25.32}&\underline{20.71}& \underline{25.57} &\underline{25.96}\\
    EditGuard$^\ddag$ \cite{editguard}  &52.92&	49.15&	48.91&	50.75&50.43\\
    CMark$^\dag$ (Ours) &\textbf{0.31}&\textbf{0.12}&\textbf{0.12}  & \textbf{0.73} &\textbf{0.32}\\
    \bottomrule[1.2pt]
    \end{tabular}}
    \label{t12}
\end{table}
\noindent\textbf{Distortions after swapping.}
As stated in the threatening Case 2, attackers attempt to interfere with face swapping detectors by introducing additional distortions to the swapped images. We simulate this scenario by applying multiple face swapping techniques and subsequently introduce normal distortions to these images. According to the results in Tab.~\ref{t6}, our approach still maintains superior detection accuracy, which significantly demonstrates the robustness of our approach even if the swapped images are further distorted. Moreover, we further compare the CMark model with the rest of watermark works. As shown in Tab.~\ref{t12}, under SimSwap \cite{simswap} technique, our approach maintains the lowest BER$^{pr}$ across various distortions, validating the strong robustness of our approach.
\begin{table}[t]
    \centering
    \footnotesize
    \renewcommand{\arraystretch}{1.2}
    \renewcommand\tabcolsep{2pt}
    \caption{Ablation under different watermark embedding locations.}
    \scalebox{0.9}{
    \begin{tabular}{c|cc|cc|cc}
    \toprule[1.2pt]
     \multirow{2}*{Location} & \multicolumn{2}{c}{SimSwap} & \multicolumn{2}{c}{ECrop}  & \multicolumn{2}{c}{ECrop$+$SimSwap} \\
     \cline{2-7}
&BER$^{pr}$$\downarrow$&ACC$\uparrow$&BER$^{pr}$$\downarrow$&BER$^{re}$$\downarrow$&BER$^{pr}$$\downarrow$&ACC$\uparrow$\\
    \hline
    All & 5.32 & 82.73 & 1.13 & 12.53 & 22.30 & 68.79\\
    Background & \textbf{0.43}&\textbf{93.01} &\underline{2.21} & \underline{13.67}& \underline{19.19} & \underline{74.83}\\
    Contour & \underline{2.60} & \underline{91.96} & \textbf{0.03} & \textbf{10.08} & \textbf{5.99} & \textbf{85.62}\\
    \bottomrule[1.2pt]
    \end{tabular}}
    \label{t9}
\end{table}

\subsection{Ablation Study}
\noindent\textbf{Embedding location.} To ensure the detection generalization, the CMark model is trained without any swapping techniques. The embedding location is thus important to ensure the watermark robustness. Tab.~\ref{t9} shows different embedding configurations. When encountering the swapping technique, the robustness is degraded after embedding into entire images as the facial regions are severely tampered. While embedding the background mitigates this impact, the robustness can not be maintained after attackers further crop the background (see threatening Case 1). Since the contour belongs to the subpart of the background and closely attaches to the face, it can be simultaneously less susceptible to malicious cropping and face swapping.

\noindent\textbf{Detection threshold.} Given the threshold overfitting when excessively adjusting $t_1$ and $t_2$ to adapt specific manipulations, we briefly let $t_1$$=$$t_2$ during the optimal threshold searching. As shown in Tab.~\ref{t10}, a lower threshold reduces detection performance as a large number of watermarked real images are predicted as either non-watermarked or watermarked fake images. Conversely, a higher threshold leads to misclassifying watermarked fake images as real ones.
\begin{table}[t]
    \centering
    \footnotesize
    \renewcommand{\arraystretch}{1}
    \renewcommand\tabcolsep{6.5pt}
    \caption{Ablation of threshold under SimSwap \cite{simswap} on CelebA-HQ \cite{celebhq} dataset.}
    \begin{tabular}{c|cc|cc}
    \toprule[1.2pt]
    \multirow{2}*{Threshold} & \multicolumn{2}{c|}{128$\times$128 }& \multicolumn{2}{c}{256$\times$256}\\
    \cline{2-5}
    &ACC$\uparrow$&F1-Score$\uparrow$&ACC$\uparrow$&F1-Score$\uparrow$\\
    \hline
    1/16 &59.66&63.83&69.17&73.41 \\
    2/16 &81.08&83.28&88.12&88.85 \\
    3/16 &\textbf{91.96}&\textbf{92.44}&\textbf{95.54}&\textbf{95.37}\\
    4/16 &\underline{90.88}&\underline{91.27}&\underline{93.36}&\underline{93.66}\\
    5/16 &86.23&88.03&90.37&91.13\\
    \bottomrule[1.2pt]
    \end{tabular}
    \label{t10}
\end{table}

\begin{figure}[t]
\centering
\includegraphics[width=\linewidth]{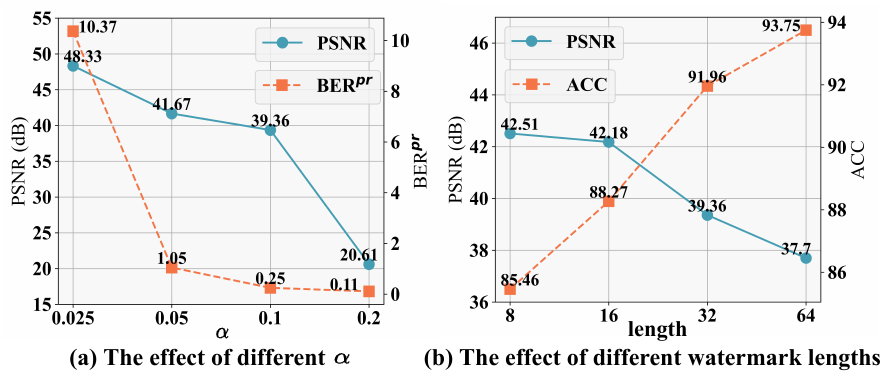}
\caption{\textbf{Ablation study of $\alpha$ and watermark length to the CMark model.} \textbf{(a)} BER$^{pr}$ is computed by average the results from the GNoise, GBlur, ECrop, and JPEG distortions. \textbf{(b)} Detection accuracy is calculated based on the swapped images from SimSwap.}
\label{s_f5}
\end{figure}

\noindent\textbf{Strength factor.} Fig.~\ref{s_f5}(a) shows the effect of varying strength factor $\alpha$. Apparently, a smaller $\alpha$ enhances watermark imperceptibility but dramatically compromises robustness against normal distortions. Conversely, a larger $\alpha$ results in a significant degradation of visual quality. An optimal balance between imperceptibility and robustness is achieved when $\alpha$=0.1. 

\noindent\textbf{Watermark length.} Fig.~\ref{s_f5}(b) depicts the influence of watermark length. It can be seen that a watermark length of 32 strikes an effective trade-off between imperceptibility and detection accuracy. A shorter length leads to low detection accuracy as the BER$_{out}$ of the real and swapped face images become less distinguishable and prone to misjudgment. On the other hand, a longer watermark length degrades visual quality due to the increased embedded information.

\noindent\textbf{Components effectiveness.} To verify the effectiveness of each component, we examine the proactive denosier (PD), the utilization of watermarked images (WI), and the contour mask constraint (MC). Results are listed in Tab. \ref{t8}. (a) PD: Contour mask consistency is important to ensure the decoding correctness. Furthermore, the extracted reference messages are significantly impacted by the normal distortions. Without the denosier, the BER$^{re}\!$ thus increases following distortion of the watermarked images. After employing the denosier, normal distorted images are standardized to a pattern closer to the clean, thereby reducing BER$^{re}$. (b) WI: There is no difference in BER$^{pr}\!$ when utilizing the encoded images as the inputs of watermark decoder. However, due to reconstruction biases in $\mathcal{E}$, the visual quality declines without fusing the original images by Eq.~\ref{e6}. (c) MC: Without the constraint in each block during training, watermark features integrate the entire image features, enhancing invisibility but leading to higher error rates after distortions.

\begin{table}[t]
    \centering
    \footnotesize
    \renewcommand{\arraystretch}{1.2}
    \renewcommand\tabcolsep{0.6pt}
    \caption{Ablation under different configurations in CMark model.}
    \scalebox{0.95}{
    \begin{tabular}{c|ccc|c|cccc}
    \toprule[1.2pt]
     Case&MC & WI & PD & Distortions & BER$^{pr}$ $\downarrow$ & BER$^{re}$ $\downarrow$ & PSNR $\uparrow$ & SSIM $\uparrow$\\
    \hline
    (a)&$\checkmark$ & $\checkmark$ & - & - &0.02 & \textbf{2.94} & 39.36& 0.971\\
    (a)&$\checkmark$ & $\checkmark$ & - & $\checkmark$ &0.27 &6.82 & 39.36& 0.971\\
    (b)&$\checkmark$ & - & $\checkmark$ & - & 0.02 & 4.69 & 37.02 & 0.946\\
    (c)&- & $\checkmark$ & $\checkmark$ & - &  \underline{0.05}&  3.98 &  39.42 & 0.977 \\
    (c)&- & $\checkmark$ & $\checkmark$ & $\checkmark$ & 0.89  & 7.38 & \textbf{39.42} & \textbf{0.977 }\\
    \hline
    &$\checkmark$ & $\checkmark$ & $\checkmark$ & - & \textbf{0.02} & \underline{3.45} &\underline{39.36} & \underline{0.972}\\
    \multirow{-2}*{Ours}&$\checkmark$ & $\checkmark$ & $\checkmark$ & $\checkmark$ & 0.25 & 5.79 &39.36 & 0.972\\
    \bottomrule[1.2pt]
    \end{tabular}}
    \label{t8}
\end{table}

\section{Limitations and Future Perspective}
\label{sec:6}
\noindent While the CMark model demonstrates promising performance in detecting face swapped images, its generalization to other face forgery types, such as face reenactment and editing, requires further investigation. Moreover, future work will also aim to enhance the model robustness against adversarial attacks, particularly in untrusted client-side deployment scenarios.

\section{Conclusion}
\label{sec:7}
\noindent To generalize the defense against different unknown face swapping techniques, we start from the face swapping purpose and propose a proactive detector that requires neither face swapping techniques for training nor large-scale message storage. CMark model focuses on the contour region surrounding the face and embeds the contour texture and face identity information into the images. Asymmetric encryption secures the model during message communication. Extensive experiments validate the generalization of face swapping detection, the visual quality of watermarked images, and the robustness of watermark across multiple scenarios.

\begin{appendices}
\section{Explanation of Encryption and Decryption}
\label{secA1}
\noindent Given a $\phi=(p-1)\cdot(q-1)$, where $p$ and $q$ are different prime numbers. A public key $e$ is randomly generated as follows:
\begin{equation}
\text{GCD}(\phi, e) = 1, \text{ s.t. } e \in (1, \phi),
\label{e14}
\end{equation}
where \text{GCD($\cdot$)} indicates the function of greatest common divisor. The private key $d$ is computed by:
\begin{equation}
d \cdot e \equiv 1 \text{ mod } \phi,
\label{e15}
\end{equation}
where $d$ mathematically represents the modular multiplicative inverse of $e$ modulo $\phi$. Therefore, the encryption function RSA$\_$En($\cdot$) \textit{w.r.t.} message \textbf{V} is expressed:
\begin{equation}
\textbf{E}_v \equiv \text{Bin}(\text{Int}(\textbf{V})^e \text{ mod } (p\cdot q)),
\label{e16}
\end{equation}
where $\text{Int}(\cdot)$ denotes the conversion of a binary number to its decimal equivalent, and $\text{Bin}(\cdot)$ refers to the conversion of a decimal number to its binary representation. During the decryption, RSA$\_$De($\cdot$) \textit{w.r.t.} \textbf{E}$_v$ is formulated as follows:
\begin{equation}
\textbf{V} \equiv \text{Bin}(\text{Int}(\textbf{E}_v)^d \text{ mod } (p\cdot q)).
\label{e17}
\end{equation}
\par When the product $p\cdot q$ is sufficiently large, reversing the $p$ and $q$ becomes computationally infeasible. This increases the difficulty of deciphering $\phi$, thereby ensuring the security of $d$, even if attackers know the $e$ and Eq.(\ref{e15}). Consequently, $\textbf{V}$ can be effectively prevented from interception.

\section{Robustness under More Image Distortions}
\label{secA3}
\begin{figure*}[t]
\centering
\includegraphics[width=15cm]{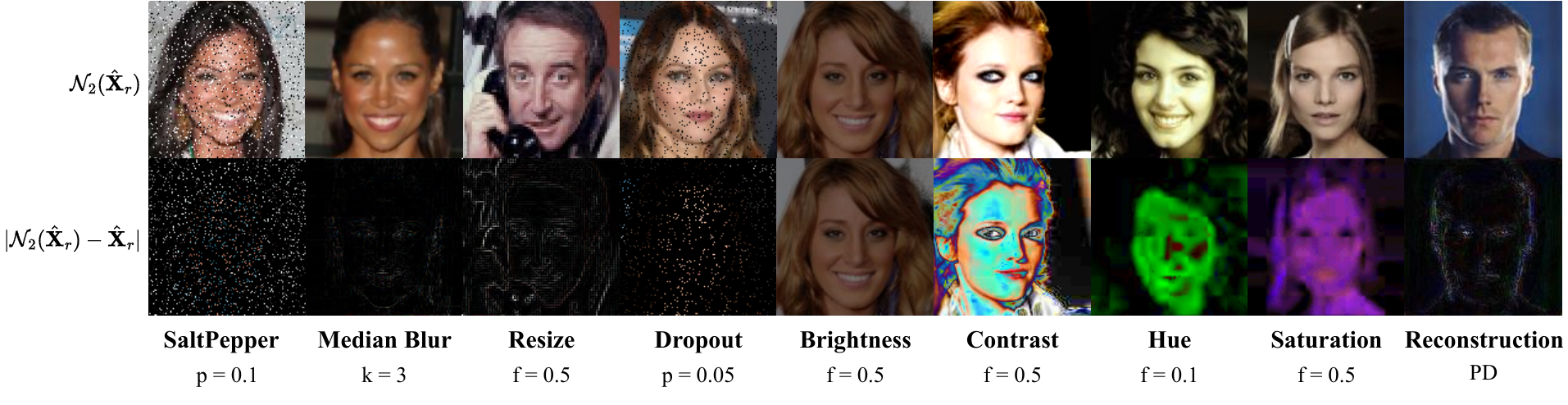}
\caption{\textbf{Qualitative results of watermarked images under different normal distortions on CelebA-HQ} \cite{celebhq}. Each column in the first row corresponds to one type of distortion. `p' denotes the distorted probability that attaches to each pixel. `k' is the kernel size, `f' represents the deviation factor related to specific distortions, and `PD' means the proposed proactive denoiser. All image sizes are equal to 128$\times$128.}
\label{s_f1}
\end{figure*}

\noindent  To validate the robustness of the CMark, we introduce more image normal distortions and compare it with other methods. The BER$^{pr}$ under these normal distortions are listed in Tab.~\ref{s_t9}. Apparently, our approach surpasses the secondary method \cite{sepmark} by over 1.24\% on average at 256$\times$256 and attains comparable results at 128$\times$128 resolution, respectively. After simulating the reconstruction distortion through the proposed denoiser, the robustness is severely degraded for FaceSigns \cite{facesigns}. However, the robustness of CMark is still maintained with few fluctuations, validating the robustness of our approach in challenging scenarios. Additional distorted watermarked instances are presented in Fig.~\ref{s_f1}.

\begin{table*}[t]
    \centering
    \footnotesize
    \renewcommand{\arraystretch}{1}
    \renewcommand\tabcolsep{1pt}
    \caption{Comparisons of BER$^{pr}$ on CelebA-HQ. The best and secondary are denoted as \textbf{bold} and \underline{underline}. `GNoise', `GBlur', `ECrop', `MBlur', and `Rec.' mean GaussianNoise, GaussianBlur, EdgeCrop, MedianBlur, and reconstruction. $\dag$ and $\ddag$ mean the model is tested on 256$\times$256 and 512$\times$512, respectively.}
    \scalebox{0.85}{
    \begin{tabular}{c|ccccccccccccc|c}
    \toprule[1.5pt]
    Method & GNoise&GBlur&ECrop&JPEG&SaltPepper & MBlur & Resize  & Dropout  & Brightness & Contrast & Hue & Saturation & Rec. & Avg. $\downarrow$\\
    \hline
    HIDDEN \cite{zhu2018hidden}&48.64&26.96&12.21&32.16&23.13&17.53&18.29&13.21&11.28&11.38&13.79&11.16&19.21&19.91\\
    PIMOG \cite{pimog}  &12.73&0.12&1.74&19.56&2.58&0.07&0.05&0.64&1.21&0.68&0.14&0.07&0.13&3.05\\
    ARWGAN \cite{arwgan} &46.55&14.78&2.37&42.58&21.13&1.14&7.51&5.15&1.02&0.91&2.94&0.86&9.08&12.00 \\
    SepMark \cite{sepmark}   &0.82&0.01&0.01&0.24&0.03&0.01&0.01&0.01&0.01&0.01&0.01&0.01&0.01&{\textbf{{0.09}}}\\
    CMark&0.47&0.03&0.03&0.46&0.08&0.04&0.13&0.02&0.02&0.06&0.03&0.02&0.30&\underline{0.13}\\
    \hline
    MBRS$^{\dag}$ \cite{mbrs}  &41.68&27.95&19.33&0.31&30.85&2.56&2.26&7.99&0.92&1.13&0.01&0.01&0.02&10.38\\
    FaceSigns$^{\dag}$ \cite{facesigns} &0.86&0.17&0.27&0.85&12.75&0.11&1.02&1.85&8.41&0.02&0.76&0.05&13.54&3.12\\
    SepMark$^{\dag}$ \cite{sepmark}   &0.06&0.01&18.13&0.01&0.01&0.01	&0.01	&0.01	&0.01	&0.01	&0.01	&0.01&0.01	&{\underline{1.41}}\\
    EditGuard$^{\ddag}$ \cite{editguard}  &15.81&98.31&3.94&23.62&3.68&93.59&49.88&0.84&0.32&0.39&4.15&0.55&51.47&26.65\\
    CMark$^{\dag}$&0.31&0.12&0.12&0.73&0.01&0.12&0.13&0.13&0.19&0.21&0.01&0.11&0.14&{\textbf{0.17}}\\
    \bottomrule[1.5pt]
    \end{tabular}}
    \label{s_t9}
\end{table*}

\begin{figure*}[t]
\centering
\includegraphics[width=13cm]{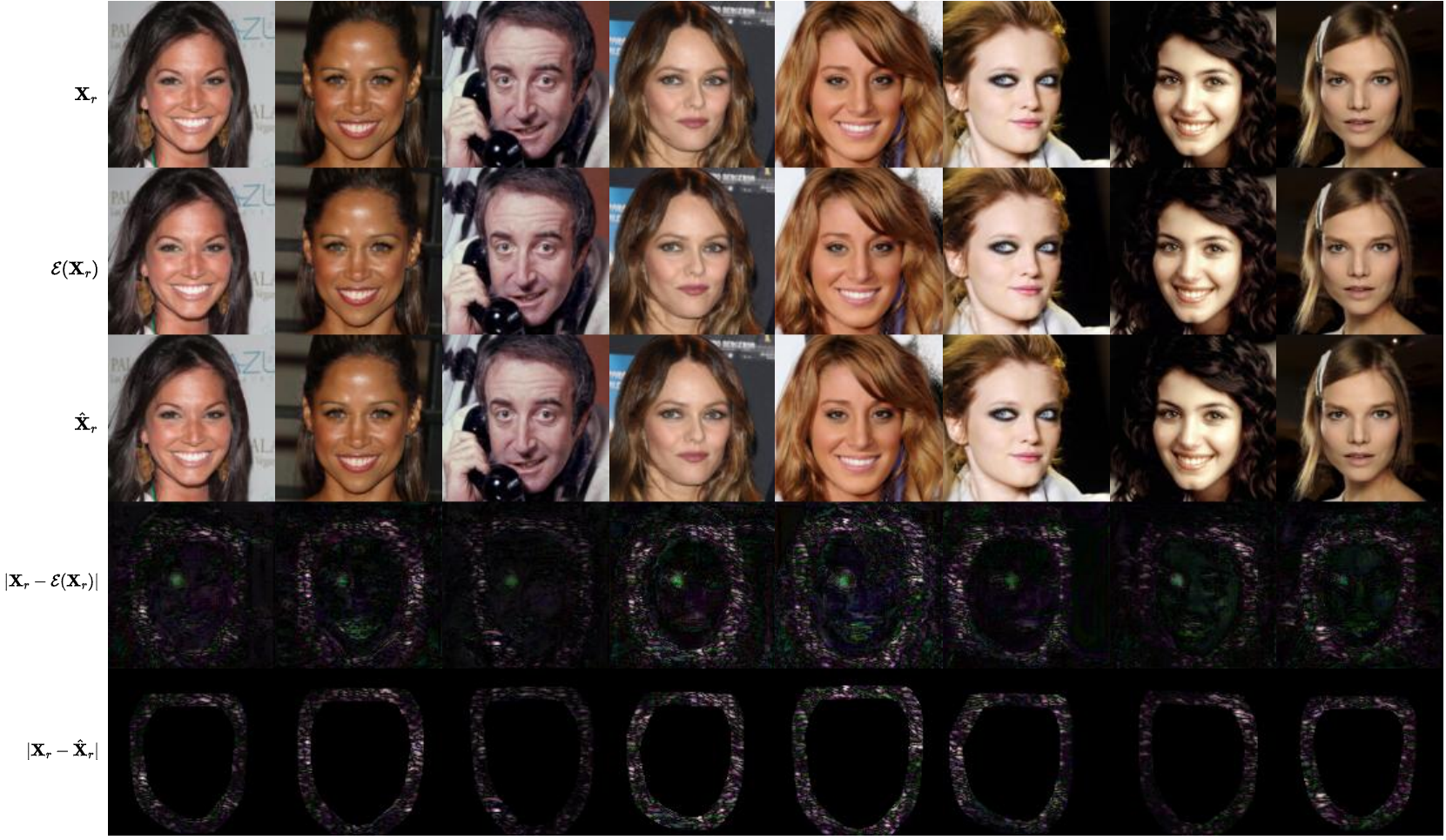}
\caption{Qualitative results on CelebA-HQ \cite{celebhq} dataset. All image sizes are equal to 128$\times$128.}
\label{s_f2}
\end{figure*}
\begin{figure*}[t]
\centering
\includegraphics[width=13.5cm]{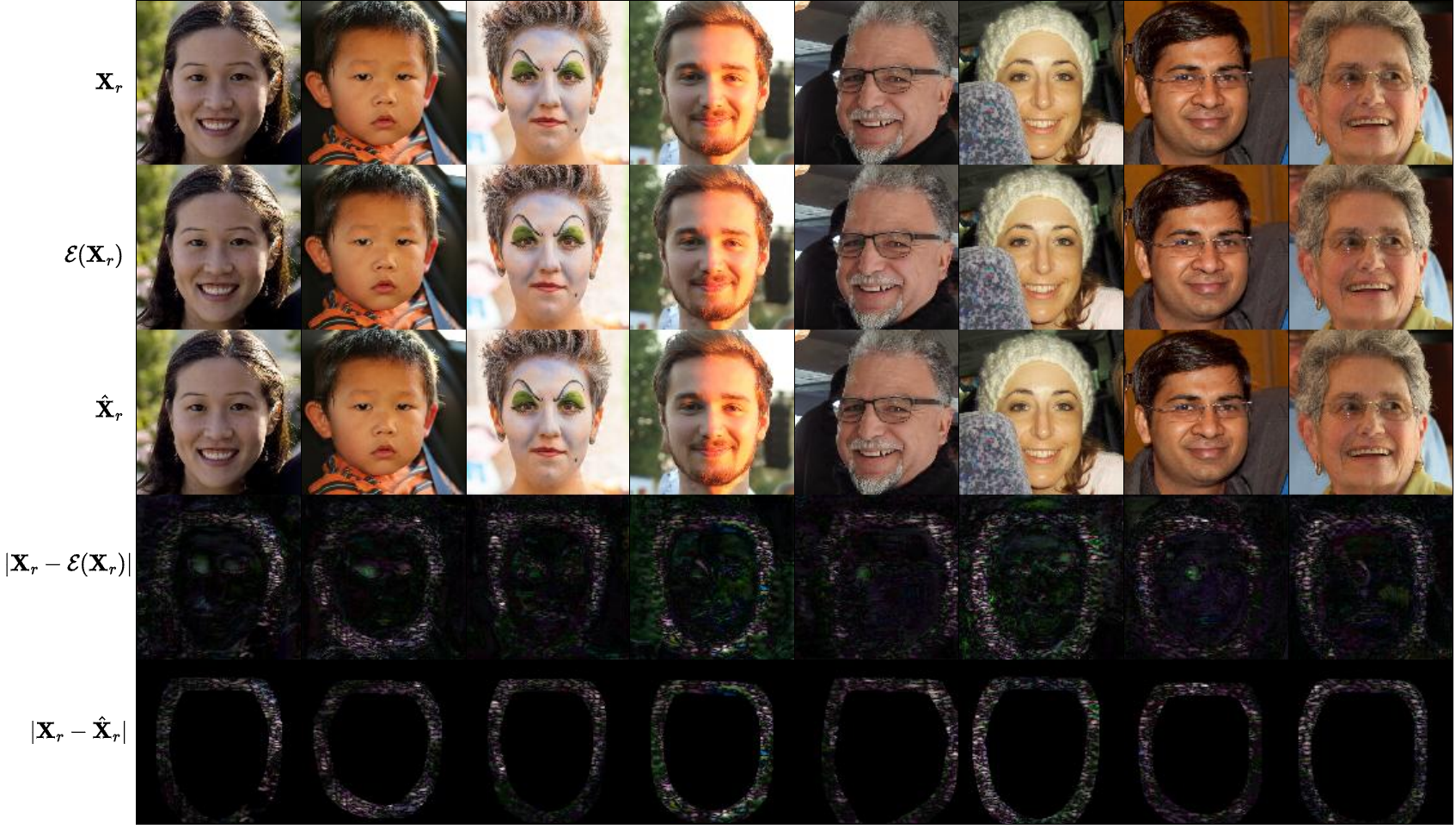}
\caption{Qualitative results on FFHQ \cite{stylegan} dataset. All image sizes are equal to 128$\times$128.}
\label{s_f3}
\end{figure*}
\section{More Illustrations of Watermarked Images}
\label{secA2}
\noindent Fig.~\ref{s_f2} illustrates more watermarked images on CelebA-HQ \cite{celebhq}. While the discrepancies between the encoded and original images are primarily located in the face contour, extra discrepancies can also be observed in the background or internal facial regions. These extra discrepancies in the encoded images arise from the reconstruction biases introduced by the watermark encoder. Accordingly, applying Eq.~\ref{e6} is important to mitigate these reconstruction biases, thereby improving the visual quality while maintaining the watermark robustness. Furthermore, the qualitative results on FFHQ \cite{stylegan} are presented in Fig.~\ref{s_f3}. The visual discrepancies between the original images and the corresponding watermarked images remain indistinguishable to human eyes, ensuring the imperceptibility of the embedded watermark.
\end{appendices}

% \noindent \textbf{Acknowledgements:} 
% This work was supported in part by NSFC under Grant NO.62176198, 62036007 and U22A2096, 
% in part by the Key Laboratory of Big Data Intelligent Computing under Grant BDIC-2023-A-004,
% in part by the Key R\&D Program of Shaanxi Province under Grant 2024GX-YBXM.135,
% in part by the  Shaanxi Province Core Technology Research and Development Project under grant 2024QY2-GJHX-11,
% in part by the Fundamental Research Funds for the Central Universities under GrantQTZX23042.

%\begin{figure}
%  \includegraphics{example.eps}
%\end{figure}
%\begin{figure*}
%  \includegraphics[width=0.75\textwidth]{example.eps}
%\end{figure*}
%

%\begin{acknowledgements}
%\end{acknowledgements}

\bibliographystyle{spbasic}
\bibliography{sn-bibliography}    % name your BibTeX data base

\begin{thebibliography}{58}
\providecommand{\natexlab}[1]{#1}
\providecommand{\url}[1]{{#1}}
\providecommand{\urlprefix}{URL }
\expandafter\ifx\csname urlstyle\endcsname\relax
  \providecommand{\doi}[1]{DOI~\discretionary{}{}{}#1}\else
  \providecommand{\doi}{DOI~\discretionary{}{}{}\begingroup \urlstyle{rm}\Url}\fi
\providecommand{\eprint}[2][]{\url{#2}}

\bibitem[{Abdi and Williams(2010)}]{pca}
Abdi H, Williams LJ (2010) Principal component analysis. Wiley interdisciplinary reviews: computational statistics 2(4):433--459

\bibitem[{Cao et~al(2022)Cao, Ma, Yao, Chen, Ding, and Yang}]{recce}
Cao J, Ma C, Yao T, Chen S, Ding S, Yang X (2022) End-to-end reconstruction-classification learning for face forgery detection. In: Proceedings of the IEEE/CVF Conference on Computer Vision and Pattern Recognition (CVPR), pp 4113--4122

\bibitem[{Chen et~al(2020)Chen, Chen, Ni, and Ge}]{simswap}
Chen R, Chen X, Ni B, Ge Y (2020) Simswap: An efficient framework for high fidelity face swapping. In: Proceedings of the 28th ACM international conference on multimedia, pp 2003--2011

\bibitem[{Cheng et~al(2024)Cheng, Yan, Zhang, Luo, Wang, and Li}]{prodet}
Cheng J, Yan Z, Zhang Y, Luo Y, Wang Z, Li C (2024) Can we leave deepfake data behind in training deepfake detector? arXiv preprint arXiv:240817052

\bibitem[{Deng et~al(2019)Deng, Guo, Xue, and Zafeiriou}]{arcface}
Deng J, Guo J, Xue N, Zafeiriou S (2019) Arcface: Additive angular margin loss for deep face recognition. In: Proceedings of the IEEE/CVF conference on computer vision and pattern recognition, pp 4690--4699

\bibitem[{Dong et~al(2023)Dong, Wang, Ji, Liang, Fan, and Ge}]{caddm}
Dong S, Wang J, Ji R, Liang J, Fan H, Ge Z (2023) Implicit identity leakage: The stumbling block to improving deepfake detection generalization. In: Proceedings of the IEEE/CVF Conference on Computer Vision and Pattern Recognition, pp 3994--4004

\bibitem[{Fang et~al(2022)Fang, Jia, Ma, Chang, and Zhang}]{pimog}
Fang H, Jia Z, Ma Z, Chang EC, Zhang W (2022) Pimog: An effective screen-shooting noise-layer simulation for deep-learning-based watermarking network. In: Proceedings of the 30th ACM international conference on multimedia, pp 2267--2275

\bibitem[{Heusel et~al(2017)Heusel, Ramsauer, Unterthiner, Nessler, and Hochreiter}]{fid}
Heusel M, Ramsauer H, Unterthiner T, Nessler B, Hochreiter S (2017) Gans trained by a two time-scale update rule converge to a local nash equilibrium. Advances in neural information processing systems 30

\bibitem[{Ho et~al(2020)Ho, Jain, and Abbeel}]{ddpm}
Ho J, Jain A, Abbeel P (2020) Denoising diffusion probabilistic models. Advances in neural information processing systems 33:6840--6851

\bibitem[{Hu et~al(2018)Hu, Shen, and Sun}]{se}
Hu J, Shen L, Sun G (2018) Squeeze-and-excitation networks. In: Proceedings of the IEEE conference on computer vision and pattern recognition, pp 7132--7141

\bibitem[{Huang et~al(2023{\natexlab{a}})Huang, Wang, Yang, Ai, Zou, Wang, and Ye}]{fsd}
Huang B, Wang Z, Yang J, Ai J, Zou Q, Wang Q, Ye D (2023{\natexlab{a}}) Implicit identity driven deepfake face swapping detection. In: Proceedings of the IEEE/CVF conference on computer vision and pattern recognition, pp 4490--4499

\bibitem[{Huang et~al(2023{\natexlab{b}})Huang, Luo, Li, Yang, Xu, and Chang}]{arwgan}
Huang J, Luo T, Li L, Yang G, Xu H, Chang CC (2023{\natexlab{b}}) Arwgan: Attention-guided robust image watermarking model based on gan. IEEE Transactions on Instrumentation and Measurement 72:1--17

\bibitem[{iProov(2024)}]{report}
iProov (2024) New threat intelligence report exposes the impact of generative ai on remote identity verification. https://www.iproov.com/press/new-threat-intelligence-report-exposes-impact-generative-ai-remote-identity-verification

\bibitem[{Jia et~al(2021)Jia, Fang, and Zhang}]{mbrs}
Jia Z, Fang H, Zhang W (2021) Mbrs: Enhancing robustness of dnn-based watermarking by mini-batch of real and simulated jpeg compression. In: Proceedings of the 29th ACM international conference on multimedia, pp 41--49

\bibitem[{Karras et~al(2017)Karras, Aila, Laine, and Lehtinen}]{progan}
Karras T, Aila T, Laine S, Lehtinen J (2017) Progressive growing of gans for improved quality, stability, and variation. CoRR abs/1710.10196, \urlprefix\url{http://arxiv.org/abs/1710.10196}, \eprint{1710.10196}

\bibitem[{Karras et~al(2019)Karras, Laine, and Aila}]{stylegan}
Karras T, Laine S, Aila T (2019) A style-based generator architecture for generative adversarial networks. In: Proceedings of the IEEE/CVF conference on computer vision and pattern recognition, pp 4401--4410

\bibitem[{Kaur et~al(2022)Kaur, Uslu, Rittichier, and Durresi}]{trust}
Kaur D, Uslu S, Rittichier KJ, Durresi A (2022) Trustworthy artificial intelligence: a review. ACM computing surveys (CSUR) 55(2):1--38

\bibitem[{Kim et~al(2022)Kim, Kim, Cho, Seo, Nam, Lee, Kim, and Lee}]{diffface}
Kim K, Kim Y, Cho S, Seo J, Nam J, Lee K, Kim S, Lee K (2022) Diffface: Diffusion-based face swapping with facial guidance. arXiv preprint arXiv:221213344

\bibitem[{Kingma and Ba(2015)}]{adam}
Kingma DP, Ba J (2015) Adam: {A} method for stochastic optimization. In: 3rd International Conference on Learning Representations, {ICLR}

\bibitem[{Li et~al(2023)Li, Qi, Liu, Di, Liu, Pei, Yi, and Zhou}]{trust2}
Li B, Qi P, Liu B, Di S, Liu J, Pei J, Yi J, Zhou B (2023) Trustworthy ai: From principles to practices. ACM Computing Surveys 55(9):1--46

\bibitem[{Li et~al(2022)Li, Wang, Yang, Wang, and Gao}]{expression1}
Li H, Wang N, Yang X, Wang X, Gao X (2022) Towards semi-supervised deep facial expression recognition with an adaptive confidence margin. In: Proceedings of the IEEE/CVF conference on computer vision and pattern recognition, pp 4166--4175

\bibitem[{Li et~al(2019)Li, Bao, Yang, Chen, and Wen}]{faceshifter}
Li L, Bao J, Yang H, Chen D, Wen F (2019) Faceshifter: Towards high fidelity and occlusion aware face swapping. arXiv preprint arXiv:191213457

\bibitem[{Li et~al(2020)Li, Bao, Zhang, Yang, Chen, Wen, and Guo}]{xray}
Li L, Bao J, Zhang T, Yang H, Chen D, Wen F, Guo B (2020) Face x-ray for more general face forgery detection. In: Proceedings of the IEEE/CVF conference on computer vision and pattern recognition, pp 5001--5010

\bibitem[{Liu et~al(2025{\natexlab{a}})Liu, Cheng, Wang, Luo, and Xu}]{liu}
Liu MH, Cheng H, Wang T, Luo X, Xu XS (2025{\natexlab{a}}) Learning real facial concepts for independent deepfake detection. arXiv preprint arXiv:250504460

\bibitem[{Liu et~al(2025{\natexlab{b}})Liu, Liu, Luo, and Xu}]{DATA}
Liu MH, Liu XQ, Luo X, Xu XS (2025{\natexlab{b}}) Data: Multi-disentanglement based contrastive learning for open-world semi-supervised deepfake attribution. arXiv preprint arXiv:250504384

\bibitem[{Liu et~al(2023)Liu, Li, Zhang, Wang, Zhang, Wang, and Nie}]{e4s}
Liu Z, Li M, Zhang Y, Wang C, Zhang Q, Wang J, Nie Y (2023) Fine-grained face swapping via regional gan inversion. In: Proceedings of the IEEE/CVF conference on computer vision and pattern recognition, pp 8578--8587

\bibitem[{Mittal et~al(2012)Mittal, Soundararajan, and Bovik}]{niqe}
Mittal A, Soundararajan R, Bovik AC (2012) Making a “completely blind” image quality analyzer. IEEE Signal processing letters 20(3):209--212

\bibitem[{Neekhara et~al(2022)Neekhara, Hussain, Zhang, Huang, McAuley, and Koushanfar}]{facesigns}
Neekhara P, Hussain S, Zhang X, Huang K, McAuley J, Koushanfar F (2022) Facesigns: semi-fragile neural watermarks for media authentication and countering deepfakes. arXiv preprint arXiv:220401960

\bibitem[{Nichol and Dhariwal(2021)}]{iddpm}
Nichol AQ, Dhariwal P (2021) Improved denoising diffusion probabilistic models. In: International conference on machine learning, PMLR, pp 8162--8171

\bibitem[{Qian et~al(2020)Qian, Yin, Sheng, Chen, and Shao}]{f3}
Qian Y, Yin G, Sheng L, Chen Z, Shao J (2020) Thinking in frequency: Face forgery detection by mining frequency-aware clues. In: European conference on computer vision, Springer, pp 86--103

\bibitem[{Rivest(1978)}]{rsa}
Rivest R (1978) A method for obtaining digital signatures and public-key cryptosystems. Communications of the ACM

\bibitem[{Rombach et~al(2021)Rombach, Blattmann, Lorenz, Esser, and Ommer}]{ldm}
Rombach R, Blattmann A, Lorenz D, Esser P, Ommer B (2021) High-resolution image synthesis with latent diffusion models. \eprint{2112.10752}

\bibitem[{Rossler et~al(2019)Rossler, Cozzolino, Verdoliva, Riess, Thies, and Nie{\ss}ner}]{ff++}
Rossler A, Cozzolino D, Verdoliva L, Riess C, Thies J, Nie{\ss}ner M (2019) Faceforensics++: Learning to detect manipulated facial images. In: Proceedings of the IEEE/CVF international conference on computer vision, pp 1--11

\bibitem[{Shiohara and Yamasaki(2022)}]{sbi}
Shiohara K, Yamasaki T (2022) Detecting deepfakes with self-blended images. In: Proceedings of the IEEE/CVF Conference on Computer Vision and Pattern Recognition, pp 18720--18729

\bibitem[{Song et~al(2020)Song, Meng, and Ermon}]{ddim}
Song J, Meng C, Ermon S (2020) Denoising diffusion implicit models. arXiv preprint arXiv:201002502

\bibitem[{Szegedy et~al(2016)Szegedy, Vanhoucke, Ioffe, Shlens, and Wojna}]{inception}
Szegedy C, Vanhoucke V, Ioffe S, Shlens J, Wojna Z (2016) Rethinking the inception architecture for computer vision. In: Proceedings of the IEEE conference on computer vision and pattern recognition, pp 2818--2826

\bibitem[{Tao et~al(2025)Tao, Le, Tan, Liu, Qin, and Zhao}]{oddn}
Tao R, Le M, Tan C, Liu H, Qin H, Zhao Y (2025) Oddn: Addressing unpaired data challenges in open-world deepfake detection on online social networks. In: Proceedings of the AAAI Conference on Artificial Intelligence, vol~39, pp 799--807

\bibitem[{Walczyna and Piotrowski(2023)}]{faceswap_survey}
Walczyna T, Piotrowski Z (2023) Quick overview of face swap deep fakes. Applied Sciences 13(11):6711

\bibitem[{Wang et~al(2021)Wang, Juefei-Xu, Luo, Liu, and Wang}]{faketagger}
Wang R, Juefei-Xu F, Luo M, Liu Y, Wang L (2021) Faketagger: Robust safeguards against deepfake dissemination via provenance tracking. In: Proceedings of the 29th ACM international conference on multimedia, pp 3546--3555

\bibitem[{Wang et~al(2023)Wang, Huang, Cheng, Ma, and Wang}]{robust}
Wang T, Huang M, Cheng H, Ma B, Wang Y (2023) Robust identity perceptual watermark against deepfake face swapping. arXiv preprint arXiv:231101357

\bibitem[{Wang et~al(2024{\natexlab{a}})Wang, Huang, Cheng, Zhang, and Shen}]{lampmark}
Wang T, Huang M, Cheng H, Zhang X, Shen Z (2024{\natexlab{a}}) Lampmark: Proactive deepfake detection via training-free landmark perceptual watermarks. In: Proceedings of the 32nd ACM International Conference on Multimedia, pp 10515--10524

\bibitem[{Wang et~al(2024{\natexlab{b}})Wang, Liao, Chow, Lin, and Wang}]{ddsurvey}
Wang T, Liao X, Chow KP, Lin X, Wang Y (2024{\natexlab{b}}) Deepfake detection: A comprehensive survey from the reliability perspective. ACM Computing Surveys

\bibitem[{Wu et~al(2020)Wu, Xiang, Gabriben, and Nicezm}]{faceswap}
Wu H, Xiang S, Gabriben, Nicezm (2020) Faceswap. \url{https://github.com/wuhuikai/FaceSwap}

\bibitem[{Wu et~al(2023{\natexlab{a}})Wu, Gan, Chen, Wan, and Lin}]{aigc}
Wu J, Gan W, Chen Z, Wan S, Lin H (2023{\natexlab{a}}) Ai-generated content (aigc): A survey. arXiv preprint arXiv:230406632

\bibitem[{Wu et~al(2023{\natexlab{b}})Wu, Liao, and Ou}]{sepmark}
Wu X, Liao X, Ou B (2023{\natexlab{b}}) Sepmark: Deep separable watermarking for unified source tracing and deepfake detection. In: Proceedings of the 31st ACM International Conference on Multimedia

\bibitem[{Xia et~al(2024)Xia, Zhou, Liu, Yuan, Wang, Li, Wang, and Gao}]{fpg}
Xia R, Zhou D, Liu D, Yuan L, Wang S, Li J, Wang N, Gao X (2024) Advancing generalized deepfake detector with forgery perception guidance. In: Proceedings of the 32nd ACM International Conference on Multimedia, pp 6676--6685

\bibitem[{Xu et~al(2025)Xu, Li, Zhu, Wang, and Gao}]{face_attribute}
Xu Y, Li H, Zhu M, Wang N, Gao X (2025) Boosting semi-supervised facial attribute recognition with dynamic threshold pairs. IEEE Transactions on Circuits and Systems for Video Technology

\bibitem[{Xu et~al(2022)Xu, Hong, Ding, Zhu, Han, Liu, and Ding}]{mobile}
Xu Z, Hong Z, Ding C, Zhu Z, Han J, Liu J, Ding E (2022) Mobilefaceswap: A lightweight framework for video face swapping. In: Proceedings of the AAAI Conference on Artificial Intelligence, vol~36, pp 2973--2981

\bibitem[{Yan et~al(2024)Yan, Zhao, Chen, Guo, Fu, Yao, Ding, and Yuan}]{time}
Yan Z, Zhao Y, Chen S, Guo M, Fu X, Yao T, Ding S, Yuan L (2024) Generalizing deepfake video detection with plug-and-play: Video-level blending and spatiotemporal adapter tuning. arXiv preprint arXiv:240817065

\bibitem[{Zhang et~al(2018)Zhang, Isola, Efros, Shechtman, and Wang}]{lpips}
Zhang R, Isola P, Efros AA, Shechtman E, Wang O (2018) The unreasonable effectiveness of deep features as a perceptual metric. In: Proceedings of the IEEE conference on computer vision and pattern recognition, pp 586--595

\bibitem[{Zhang et~al(2024{\natexlab{a}})Zhang, Li, Yu, Xu, Li, and Zhang}]{editguard}
Zhang X, Li R, Yu J, Xu Y, Li W, Zhang J (2024{\natexlab{a}}) Editguard: Versatile image watermarking for tamper localization and copyright protection. In: Proceedings of the IEEE/CVF conference on computer vision and pattern recognition, pp 11964--11974

\bibitem[{Zhang et~al(2024{\natexlab{b}})Zhang, Ye, Xie, Tang, Liao, Liu, Chen, and Deng}]{dual}
Zhang Y, Ye D, Xie C, Tang L, Liao X, Liu Z, Chen C, Deng J (2024{\natexlab{b}}) Dual defense: Adversarial, traceable, and invisible robust watermarking against face swapping. IEEE Transactions on Information Forensics and Security

\bibitem[{Zhao et~al(2023{\natexlab{a}})Zhao, Rao, Shi, Liu, Zhou, and Lu}]{diffswap}
Zhao W, Rao Y, Shi W, Liu Z, Zhou J, Lu J (2023{\natexlab{a}}) Diffswap: High-fidelity and controllable face swapping via 3d-aware masked diffusion. In: Proceedings of the IEEE/CVF Conference on Computer Vision and Pattern Recognition, pp 8568--8577

\bibitem[{Zhao et~al(2023{\natexlab{b}})Zhao, Liu, Ding, Liu, Zhu, and Yu}]{proactive}
Zhao Y, Liu B, Ding M, Liu B, Zhu T, Yu X (2023{\natexlab{b}}) Proactive deepfake defence via identity watermarking. In: Proceedings of the IEEE/CVF winter conference on applications of computer vision, pp 4602--4611

\bibitem[{Zhou et~al(2024)Zhou, Li, Wu, Li, Dong et~al}]{freq}
Zhou J, Li Y, Wu B, Li B, Dong J, et~al (2024) Freqblender: Enhancing deepfake detection by blending frequency knowledge. Advances in Neural Information Processing Systems 37:44965--44988

\bibitem[{Zhu et~al(2022)Zhu, Wu, Zhu, Jiang, Tang, Zhang, Liu, and Loy}]{celebhq}
Zhu H, Wu W, Zhu W, Jiang L, Tang S, Zhang L, Liu Z, Loy CC (2022) Celebv-hq: A large-scale video facial attributes dataset. In: European conference on computer vision, Springer, pp 650--667

\bibitem[{Zhu et~al(2018)Zhu, Kaplan, Johnson, and Fei-Fei}]{zhu2018hidden}
Zhu J, Kaplan R, Johnson J, Fei-Fei L (2018) Hidden: Hiding data with deep networks. In: Proceedings of the European Conference on Computer Vision (ECCV)

\bibitem[{Zhu et~al(2021)Zhu, Li, Wang, Xu, and Sun}]{megafs}
Zhu Y, Li Q, Wang J, Xu CZ, Sun Z (2021) One shot face swapping on megapixels. In: Proceedings of the IEEE/CVF conference on computer vision and pattern recognition, pp 4834--4844

\end{thebibliography}
%\begin{thebibliography}{}
% etc
%\end{thebibliography}

\end{sloppypar}
\end{document}